\documentclass[10pt,twocolumn,letterpaper]{article}

\usepackage{iccv}
\usepackage{times}
\usepackage{epsfig}
\usepackage{graphicx}
\usepackage{amsmath}
\usepackage{amssymb}

\usepackage[pagebackref=true,breaklinks=true,letterpaper=true,colorlinks,bookmarks=false]{hyperref}
\usepackage[table,xcdraw]{xcolor}
\iccvfinalcopy 

\def\iccvPaperID{735} 

\ificcvfinal\fi
\begin{document}

\title{Can Image Retrieval help Visual Saliency Detection?}

\author{Shuang Li\\
{\tt\small sli@ee.cuhk.edu.hk}
\and
Peter Mathews\\
{\tt\small peter.mathews@adelaide.edu.au}
}

\maketitle
\thispagestyle{empty}

\begin{abstract}
  We propose a novel image retrieval framework for visual saliency detection using information about salient objects contained within bounding box annotations for similar images.
  For each test image, we train a customized SVM from similar example images to predict the saliency values of its object proposals and generate an external saliency map \textbf{(ES)} by aggregating the regional scores.
  To overcome limitations caused by the size of the training dataset, we also propose an internal optimization module which computes an internal saliency map \textbf{(IS)} by measuring the low-level contrast information of the test image.
  The two maps, ES and IS, have complementary properties so we take a weighted combination to further improve the detection performance.
  Experimental results on several challenging datasets demonstrate that the proposed algorithm performs favorably against the state-of-the-art methods \footnote{This paper was an undergraduate final year project finished by Shuang Li at Dalian University of Technology in 2015 and was later revised by Peter Mathews.}.
\end{abstract}

Visual saliency detection aims to find the most distinctive or important regions in an image and often serves as a preprocessing step for many computer vision tasks. Although numerous models and algorithms have been proposed in recent years, it remains a challenging problem to find salient regions with great accuracy.
Existing methods perform saliency detection either solely based on the test image itself~\cite{achanta2009frequency,itti1998model,jiang2013saliencyours,wei2012geodesic} or train parameters from a large dataset~\cite{jiang2013salient11,jiang2013salient,liu2011learning,lu2014learning}.
The former types of approaches typically focus on low-level contrast properties so are limited when the salient object is similar in color to the background.
On the other hand, the latter supervised approaches train a generally suitable model for all test samples, which may not be the optimal solution for each specific image. Therefore, a good training set should be customized to the individual image.

%
\begin{figure}[t]
\begin{center}
\begin{tabular}{@{}c@{}c@{}c@{}c@{}c}
   \includegraphics[height=0.14\linewidth,width=0.189\linewidth]{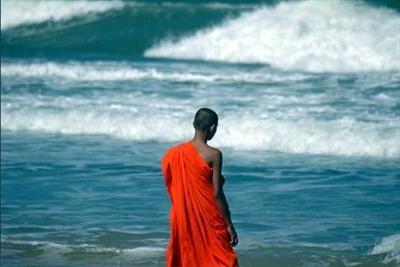}  \ &
   \includegraphics[height=0.14\linewidth,width=0.189\linewidth]{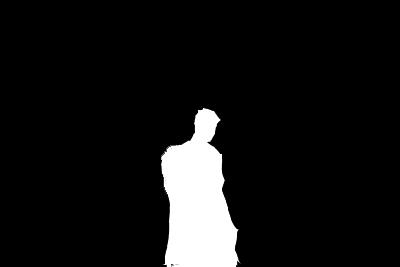}  \ &
   \includegraphics[height=0.14\linewidth,width=0.189\linewidth]{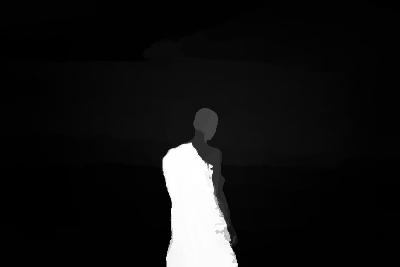}  \ &
   \includegraphics[height=0.14\linewidth,width=0.189\linewidth]{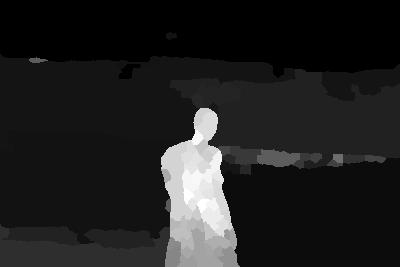}  \ &
   \includegraphics[height=0.14\linewidth,width=0.189\linewidth]{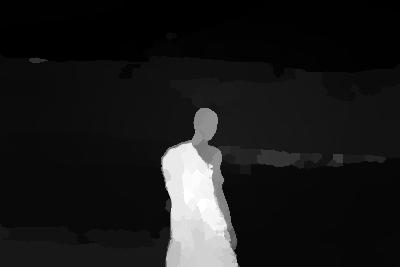}  \ \\
   \includegraphics[height=0.14\linewidth,width=0.189\linewidth]{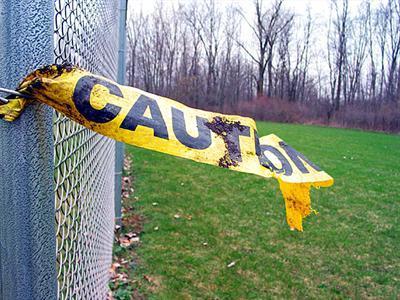}  \ &
   \includegraphics[height=0.14\linewidth,width=0.189\linewidth]{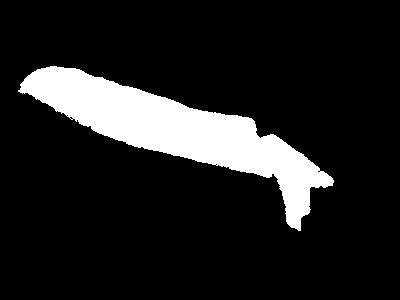}  \ &
   \includegraphics[height=0.14\linewidth,width=0.189\linewidth]{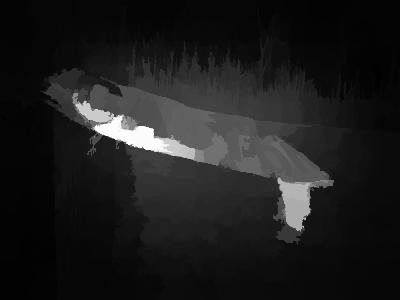}  \ &
   \includegraphics[height=0.14\linewidth,width=0.189\linewidth]{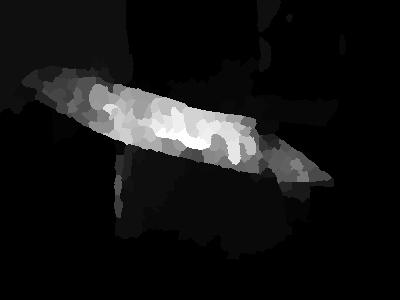}  \ &
   \includegraphics[height=0.14\linewidth,width=0.189\linewidth]{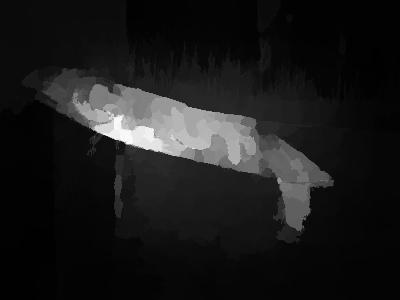}  \ \\
   \includegraphics[height=0.14\linewidth,width=0.189\linewidth]{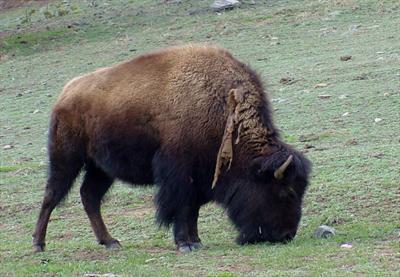}  \ &
   \includegraphics[height=0.14\linewidth,width=0.189\linewidth]{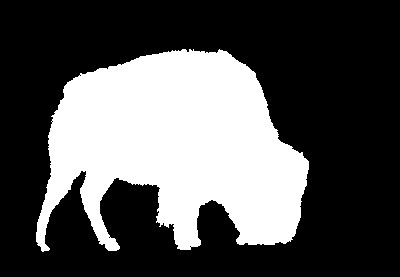}  \ &
   \includegraphics[height=0.14\linewidth,width=0.189\linewidth]{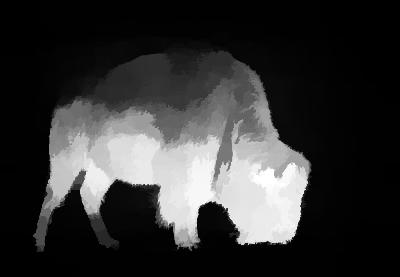}  \ &
   \includegraphics[height=0.14\linewidth,width=0.189\linewidth]{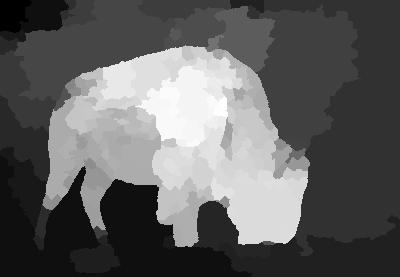}  \ &
   \includegraphics[height=0.14\linewidth,width=0.189\linewidth]{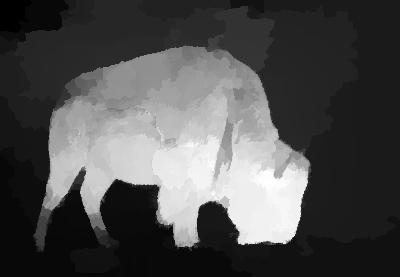}  \ \\
   \includegraphics[height=0.14\linewidth,width=0.189\linewidth]{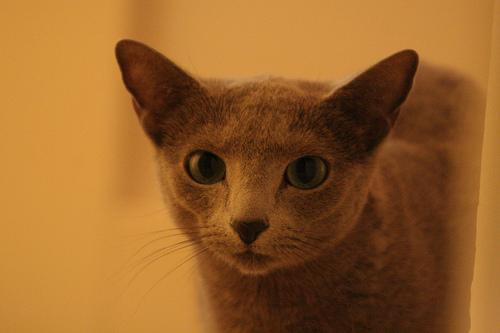}  \ &
   \includegraphics[height=0.14\linewidth,width=0.189\linewidth]{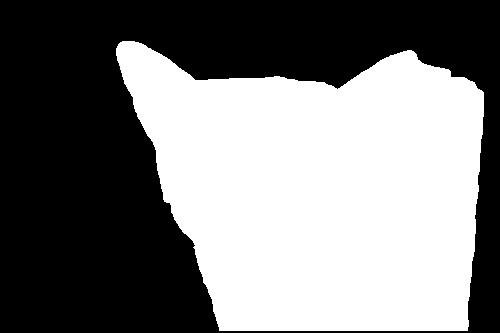}  \ &
   \includegraphics[height=0.14\linewidth,width=0.189\linewidth]{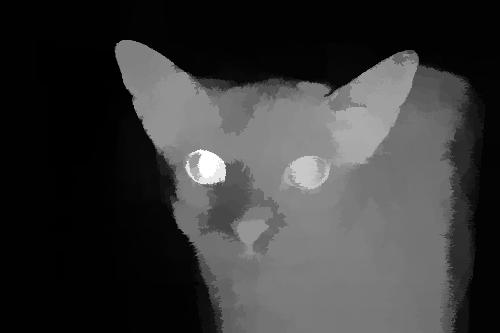}  \ &
   \includegraphics[height=0.14\linewidth,width=0.189\linewidth]{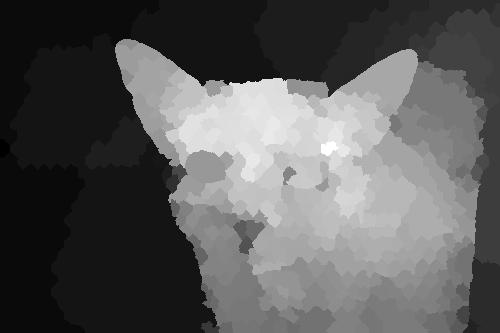}  \ &
   \includegraphics[height=0.14\linewidth,width=0.189\linewidth]{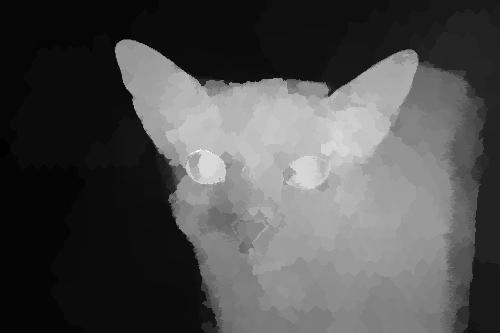}  \ \\
   {\small (a)} & {\small(b)} & {\small(c)} & {\small(d)} & {\small(e)} \ \\
\end{tabular}
\end{center}
\caption{Saliency maps generated by the proposed method:
(a) Test image.
(b) Ground truth.
(c) Internal saliency map.
(d) External saliency map.
(e) Integrated result.
\label{fig:first}}
\end{figure}
%
\begin{figure*}
\begin{center}
\includegraphics[height=0.35\linewidth]{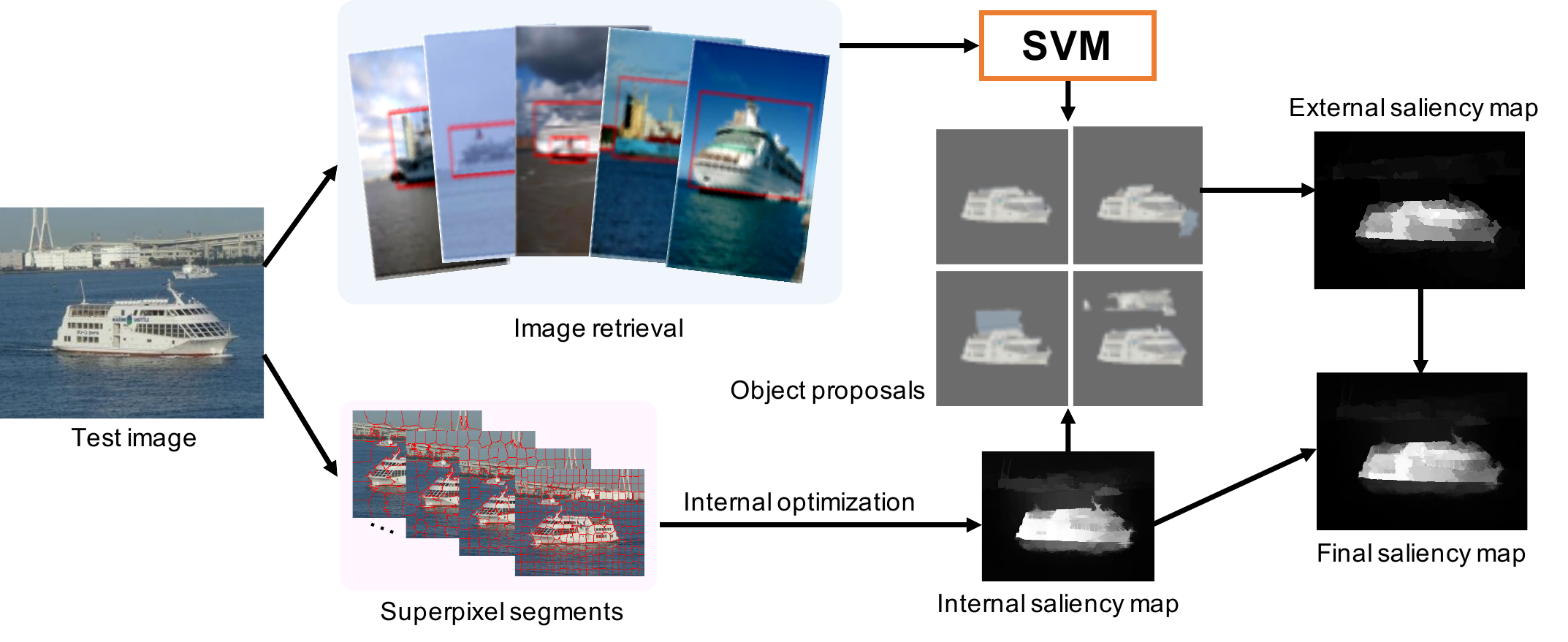}  \
\end{center}
   \caption{Pipeline of proposed algorithm.}
\label{fig:pipeline}
\end{figure*}
%
The human vision system is sensitive to intensive visual stimuli, such as color, texture and orientation.
However, the human ability to separate salient objects from chaotic backgrounds relies on prior knowledge accumulated by years of learning.
Taking the test image in Figure~\ref{fig:pipeline} for example, the salient ship and background buildings have similar appearances, but we can immediately localize the ship because a general awareness has been formed in our mind after seeing hundreds of ship models.
Similar examples can enhance our contextual knowledge and provide valuable prior information to the test image.

We address the saliency detection problem from a different perspective than previous works.
Motivated by the fact that similar images with pre-stored bounding box annotations contain important cues of shapes, positions, and colors of the target objects, we design an image retrieval approach that searches for similar example images from a large dataset and transfers saliency information from the examples to the test image.
Compared with individual image-based approaches, most of which focus on the feature contrast within the test image, our method utilizes more accurate object information and is more robust to complex backgrounds.
Furthermore, instead of using the whole dataset as training samples and treating each individual image equally, we select a subset of similar examples to train a customized SVM for each test image on-the-fly.
However, due to the finite size of the annotation dataset, some test images with rare contents may not find sufficiently similar examples.
Therefore, we also propose an internal optimization module based on low-level contrast to assist the image retrieval
and take a weighted sum to construct the final saliency map \textbf{(EIS)}.
The main contributions of our work can be summarized as follows:
\begin{itemize}
  \item We propose a novel image retrieval framework, which addresses saliency detection from a new perspective by transferring high-level object information from similar examples to the test image.
\vspace{+0.001mm}
  \item We introduce an effective internal optimization module which explores the discriminability and similarity between each pair of superpixels within the test image and serves as an essential supplement to the image retrieval.
\end{itemize}

The pipeline of proposed method is shown in Figure~\ref{fig:pipeline}.
We first search for similar examples from a large dataset and train a customized SVM classifier for each test image.
Then we generate an internal saliency map by solving a joint optimization problem.
We pick out the most salient object proposals in the test image based on the internal saliency map and predict their saliency values using an SVM classifier.
By computing the sum of saliency values, an external saliency map is constructed.
We further fuse it with the internal saliency map to generate the final saliency map.
%
%
Extensive experiments on four benchmark datasets demonstrate that the proposed algorithm outperforms most of the state-of-the-art saliency detection methods.
Several example results are shown in Figure~\ref{fig:first}.
\begin{table*}
\small
\begin{center}
\begin{tabular}{c|c||c|c||c|c|c|c|}
	\hline
	\multicolumn{2}{|c||}{\textbf{Image Features}} &
	\multicolumn{2}{c||}{\textbf{Superpixel Features}} &
	\multicolumn{4}{c|}{\textbf{Regional Features}} \\
	\hline
	\multicolumn{1}{|c|} {\textbf{features}} & \textbf{dim} & \textbf{features} & \textbf{dim} & \textbf{features} & \textbf{dim} & \textbf{features} & \textbf{dim} \\
	\hline
	\multicolumn{1}{|c|}{caffe features} & 4096 & average RGB values & 3 & RGB histogram distance & 4 & superpixel features & 30 \\
	\hline
	\multicolumn{1}{|c|}{--} & - & average Lab values & 3 & Lab histogram distance & 4 & region area & 1 \\
	\hline
	\multicolumn{1}{|c|}{--} & - & average HSV values & 3 & HSV histogram distance & 4 & max region height & 1 \\
	\hline
	\multicolumn{1}{|c|}{--} & - & absolute response of LM filters & 15 & mean RGB distance & 12 & max region height & 1 \\
	\hline
	\multicolumn{1}{|c|}{--} & - & coordinates & 15 & mean lab distance & 12 & max region height & 1 \\
	\hline
	\multicolumn{1}{|c|}{--} & - & -- & - & mean HSV distance & 12 & -- & - \\
	\hline
\end{tabular}
\end{center}
\caption{Detailed feature components of each image, superpixel and region.} \label{fig:table1}
\end{table*}
\section{Related Work}

Significant improvement in saliency detection has been witnessed in the past decade. Numerous unsupervised and supervised saliency detection methods have been proposed under different theoretical models \cite{itti1998model,li2015adaptive,zhu2014saliency,cheng2011global,lu2014learning}. However, few works address this problem from the perspective of image retrieval.

Most unsupervised algorithms are based on low-level features and perform saliency detection directly on the individual image.
Itti et al.~\cite{itti1998model} propose a saliency model which linearly combines image features including color, intensity and orientation over different scales to detect local conspicuity. However, this method tends to highlight the salient pixels and loses object information.
Zhu et al.~\cite{zhu2014saliency} propose a background measurement, boundary connectivity, to characterize the spatial layout of image regions.
In ~\cite{cheng2011global}, Cheng et al. address saliency detection based on the global region contrast, which simultaneously considers the spatial coherence across the regions and the global contrast over the entire image.
However, unsupervised algorithms lose object information and easily get affected by complex backgrounds.

Supervised methods always take a large dataset of training samples and contain high-level object information when computing saliency maps.
Liu et al.~\cite{liu2011learning} regard saliency detection as a binary labeling task and combine multi-features with a conditional random field (CRF) to generate the saliency maps.
Lu et al.~\cite{lu2014learning} search for optimal seeds by combining bottom-up saliency maps and mid-level vision cues.
However, training on a large dataset cannot ensure generating a good classifier, since it is hard to balance a large number of images with various appearances and categories.
If the training set is not large enough, the classifier becomes less robust.
Different to most supervised saliency detection methods, we train an optimal classifier for each test image by selecting training samples just from similar images instead of the whole training set.
Our image retrieval framework considers the specificity of each individual image and better designs the training set, thus generating more accurate saliency maps.

In ~\cite{marchesotti2009framework}, Marchesotti et al. also proposed to retrieve similar images for saliency detection.
However, our approach is different from theirs in three aspects.
First, we address saliency detection based on region proposals, which contain a large amount of shape and boundary information of salient regions and keep the consistency of the whole object or part of it.
Second, our approach uses a more discriminative SVM, instead of distance-based classification, to better predict the saliency values of object proposals. Our annotation database consists of 50,000 images, which is large enough to contain similar examples for most test images.
%
%
Third, unlike ~\cite{marchesotti2009framework} which relies purely on a retrieved list and thus potentially suffers from retrieval errors for uncommon objects, we use internal saliency cues with external high-level retrieved information to leverage the best out of both schemes.
Our method combines the supervised and unsupervised algorithms, considering high-level object concepts and low-level contrast simultaneously, and thus can uniformly highlight the whole salient region with explicit object boundaries and achieves better performance on the PR curves.

\section{Internal Optimization Module}

The performance of the proposed image retrieval framework relies heavily on the object proposals. Therefore, we first present a novel internal optimization module to generate a relatively accurate saliency map for the subsequent proposal generation and saliency integration.
%
%
We first decompose the image to superpixels, then jointly optimize superpixel prior, discriminability and similarity terms under a single objective function.
The superpixel prior, obtained by the sum of objectness scores within a superpixel, provides an essential saliency estimation of the test image.
The superpixel discriminability aims at identifying salient superpixels by exploring the distinctiveness between each pair.
The superpixel similarity term  tries to cluster superpixels with similar appearances together using the N-cut~\cite{shi2000normalized} algorithm.
Since multiple saliency estimations using different cues may enhance relevant information, we propose to fuse these three terms together to make the best of complementary properties to generate more accurate saliency maps.

In this section, we first introduce the superpixel features, then provide a detailed explanation of the superpixel prior, discriminability, and similarity terms. Finally, we jointly optimize an objective function to compute the internal saliency map.
\subsection{Superpixel Features}

%
Superpixel segmentation algorithms generate compact and uniform superpixels, thus greatly reducing the complexity of subsequent vision tasks.
Lacking the knowledge of size and position of objects, we produce six layers of superpixels using the SLIC algorithm with different parameters~\cite{achanta2012slic}
and construct a 30-dimensional feature vector $x_i^s\in \mathbb{R}^{30}$ that captures color, texture, and position information to describe each superpixel. The detailed feature components are summarized in Table~\ref{fig:table1}.
The color features, including RGB, Lab, and HSV, have been widely adopted by previous saliency detection methods and contribute significantly to the algorithm performance. In addition, we use the absolute response of LM filters, proposed by~\cite{jiang2013salient11}, to represent texture features and extract center and boundary coordinates as position information.
\subsection{Superpixel Prior}

In~\cite{alexe2012measuring}, Alexe et al. present an objectness approach to measure the likelihood of an image window containing an object.
We generate a pixel-wise objectness map ${\cal O}(p)$ by adding all the windows together and define the superpixel prior as ${m_i} = \sum\limits_{p \in i} {{\cal O}(p)}$, where $p$ is a pixel within superpixel $i$. The prior vector $m$ is formed by stacking $m_i$.
%
%
We aim at computing the saliency score $\ell_i$ of each superpixel, therefore constructing a linear term as follows:
\begin{equation}\label{second}
    {f_{Prior}}(\ell ) =  - {\ell ^T}m ,
\end{equation}
where $\ell$ is the score vector obtained by stacking $\ell_i$.
The prior score just provides a rough saliency estimation of each superpixel and more attention should be put on the internal structure of the test image by exploring the superpixel discriminability and similarity as described in the following sections.
\subsection{Superpixel Discriminability}

To discriminate which are the salient superpixels,
we adopt a discriminative learning approach~\cite{joulin2010discriminative} to address this problem by solving a ridge regression objective function:
%
\begin{equation}\label{third}
    \mathop {\min }\limits_{\omega ,c} \frac{1}{n}\sum\limits_{i = 1}^n {\left\| {{\ell _i} - \omega {x_i^s} - c} \right\|_2^2 + \lambda \left\| \omega  \right\|_2^2} ,
\end{equation}
where $\omega $ and $\lambda $ are the weight vector and weight parameter respectively, $n$ is the number of superpixels, and $c$ is a bias. Following~\cite{bach2008diffrac}, the objective function can be transformed to a quadratic form with a closed solution:
%
\begin{equation}\label{forth}
    {f_{Disc}}(\ell ) = {\ell ^T}{\rm U}\ell ,
\end{equation}
where ${\rm U} = \frac{1}{n}{\Pi _n}({{\rm I}_n} - X{({X^T}{\Pi _n}X + n\kappa {{\rm I}_{30}})^{ - 1}}{X^T}){\Pi _n}$,
$X \in \mathbb{R}^{n\times 30}$ is obtained by stacking $x_i^s$,
${\Pi _n}{\rm{ = }}{{\rm I}_n} - \frac{1}{n}{1_n}1_n^T$ is the centering projection matrix,
and $\kappa  = 0.01$ is a weight parameter.
The quadratic function detects salient superpixels by exploring the nonlinearity and discriminability of their features based on positive definite kernels, and assigns distinctive labels to different superpixels.
\subsection{Superpixel Similarity}

Superpixels with similar features are expected to have similar saliency values.
In this part, we construct an affinity matrix to measure the similarity of superpixels:
\begin{equation}\label{fifth}
    {W_{i,j}} = \exp \left( { - {{\sqrt {\sum\limits_{k = 1}^{30} {{{\left| {x_{i,k}^s - x_{j,k}^s} \right|}^2}} } } \mathord{\left/
    {\vphantom {{\sqrt {\sum\limits_{k = 1}^{30} {{{\left| {x_{i,k}^s - x_{j,k}^s} \right|}^2}} } } {{\sigma ^2}}}} \right.
    \kern-\nulldelimiterspace} {{\sigma ^2}}}} \right) ,
\end{equation}
where $\sigma $ is a weight parameter to control the strength of distances.
In ~\cite{shi2000normalized}, Shi and Malik propose a normalized clustering algorithm to compute the cluster labels by finding the second smallest eigenvector of the normalized Laplacian matrix ${\rm L} = {{\rm I}_n} - {D^{ - \frac{1}{2}}}W{D^{ - \frac{1}{2}}}$,
where $D$ is the diagonal matrix of $W$. However, we find the unnormalized form achieves better performance in our experiments,
with the objective function constructed as follows:
\begin{equation}\label{seventh}
    {f_{Simi}}(\ell ) = {\ell ^T}(D - W)\ell .
\end{equation}
%
In contrast to superpixel discriminability, superpixel similarity focuses on clustering superpixels together based on similarity. We construct the internal saliency map using the superpixel prior.
\begin{figure}[t]
\begin{center}
\begin{tabular}{@{}c@{}c@{}c@{}c@{}c@{}c}
   \includegraphics[height=0.13\linewidth,width=0.16\linewidth]{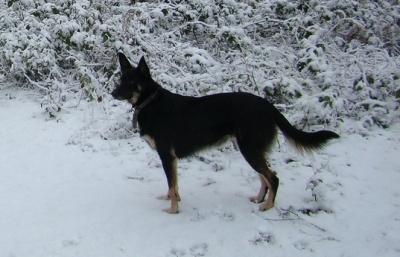}  \ &
   \includegraphics[height=0.13\linewidth,width=0.16\linewidth]{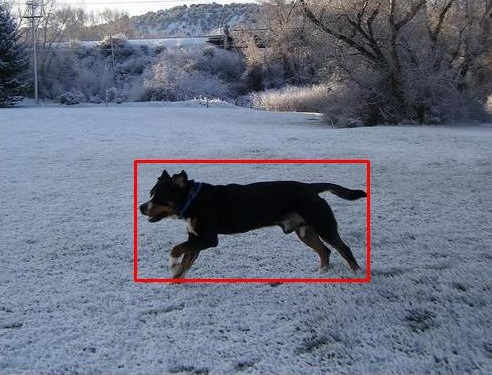}  \ &
   \includegraphics[height=0.13\linewidth,width=0.16\linewidth]{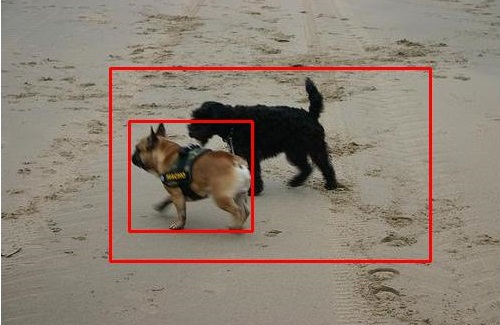}  \ &
   \includegraphics[height=0.13\linewidth,width=0.16\linewidth]{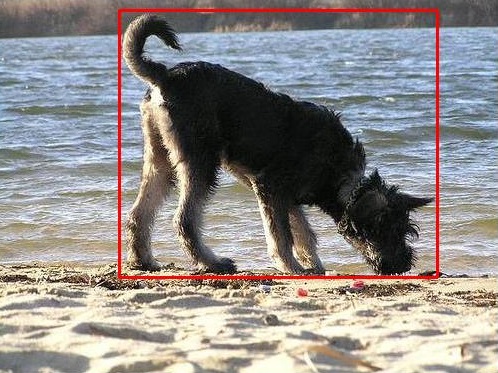}  \ &
   \includegraphics[height=0.13\linewidth,width=0.16\linewidth]{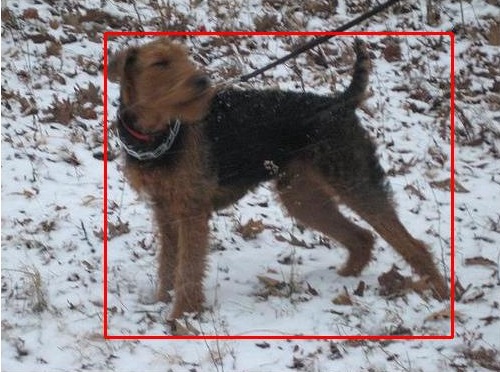}  \ &
   \includegraphics[height=0.13\linewidth,width=0.16\linewidth]{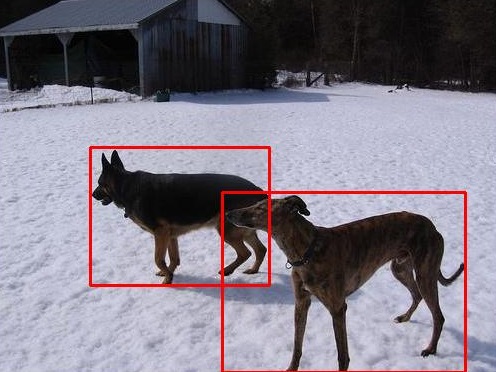}  \ \\
   \includegraphics[height=0.13\linewidth,width=0.16\linewidth]{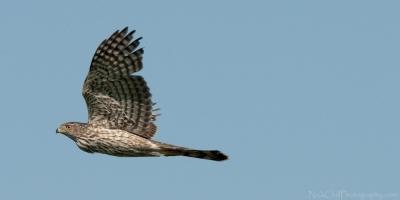}  \ &
   \includegraphics[height=0.13\linewidth,width=0.16\linewidth]{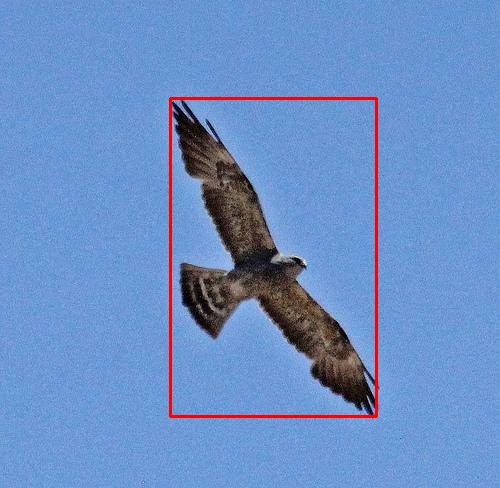}  \ &
   \includegraphics[height=0.13\linewidth,width=0.16\linewidth]{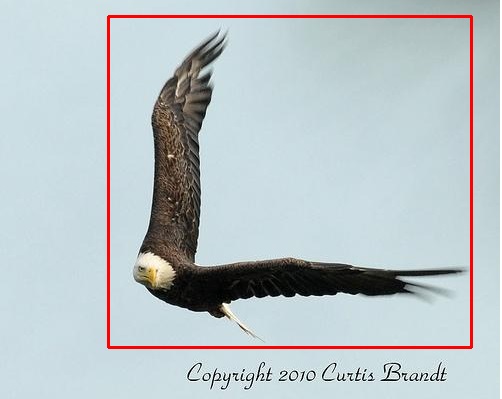}  \ &
   \includegraphics[height=0.13\linewidth,width=0.16\linewidth]{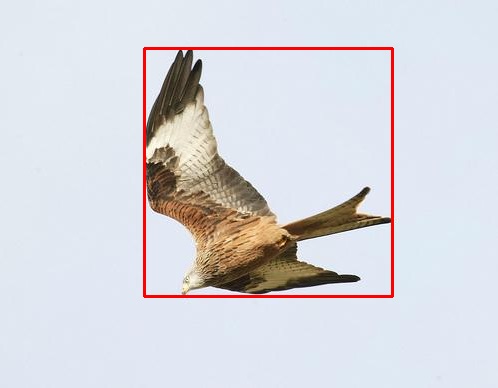}  \ &
   \includegraphics[height=0.13\linewidth,width=0.16\linewidth]{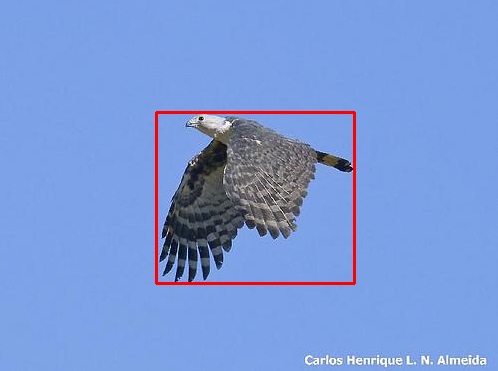}  \ &
   \includegraphics[height=0.13\linewidth,width=0.16\linewidth]{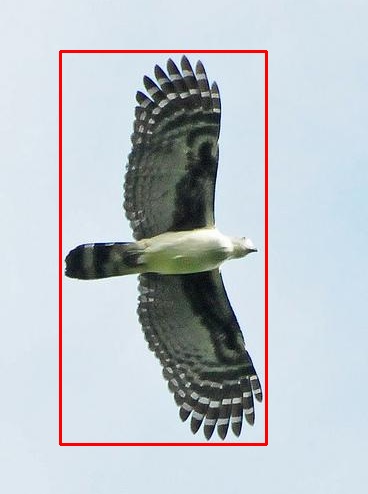}  \ \\
   \includegraphics[height=0.13\linewidth,width=0.16\linewidth]{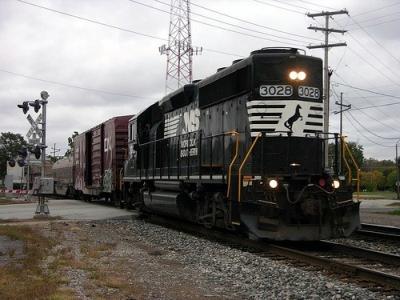}  \ &
   \includegraphics[height=0.13\linewidth,width=0.16\linewidth]{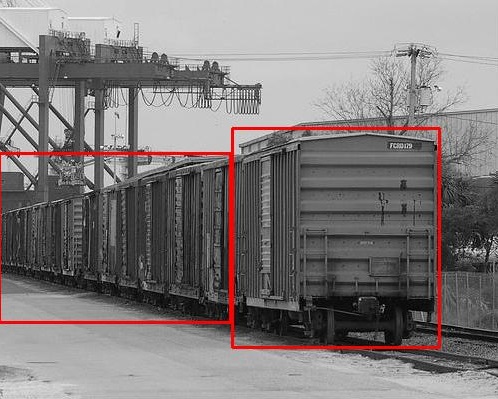}  \ &
   \includegraphics[height=0.13\linewidth,width=0.16\linewidth]{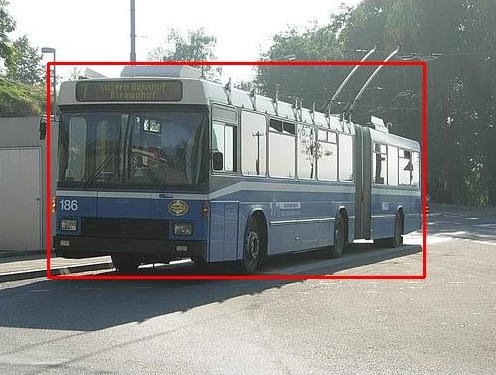}  \ &
   \includegraphics[height=0.13\linewidth,width=0.16\linewidth]{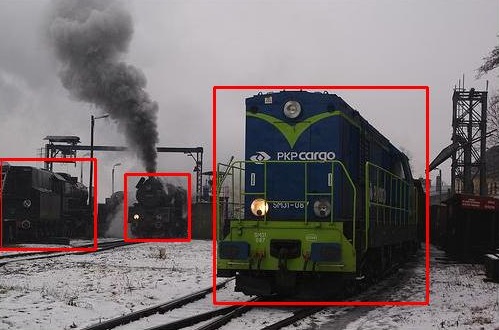}  \ &
   \includegraphics[height=0.13\linewidth,width=0.16\linewidth]{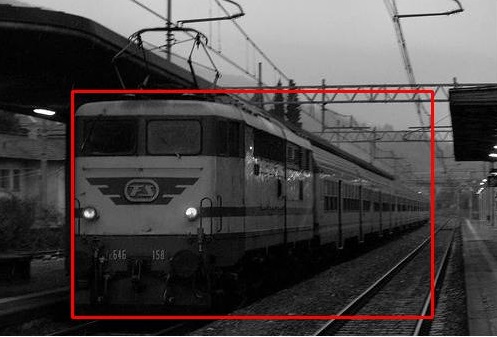}  \ &
   \includegraphics[height=0.13\linewidth,width=0.16\linewidth]{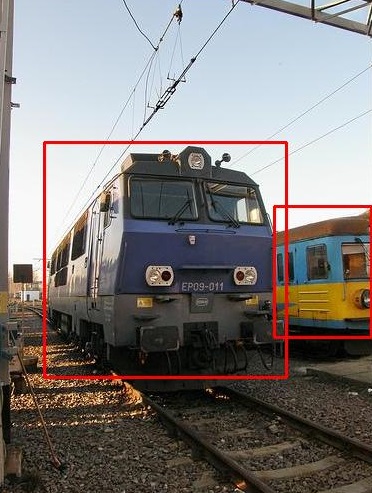}  \ \\
   %
   \includegraphics[height=0.13\linewidth,width=0.16\linewidth]{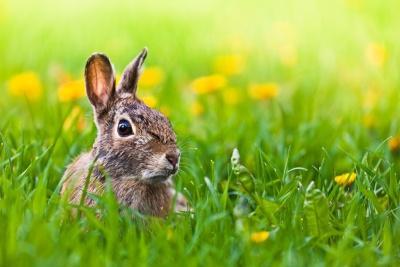}  \ &
   \includegraphics[height=0.13\linewidth,width=0.16\linewidth]{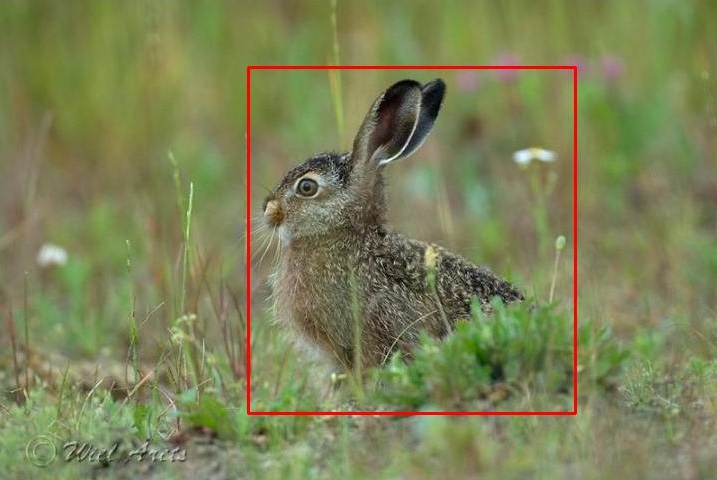}  \ &
   \includegraphics[height=0.13\linewidth,width=0.16\linewidth]{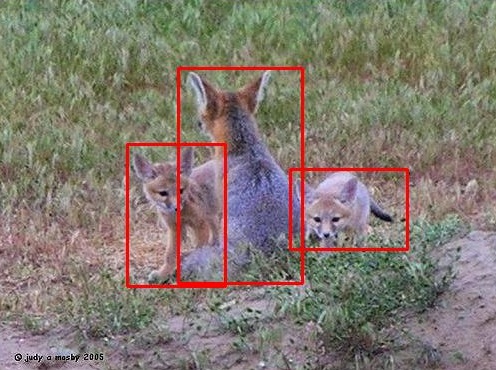}  \ &
   \includegraphics[height=0.13\linewidth,width=0.16\linewidth]{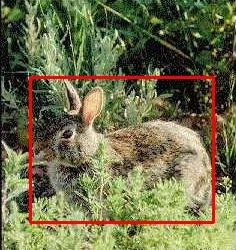}  \ &
   \includegraphics[height=0.13\linewidth,width=0.16\linewidth]{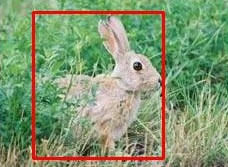}  \ &
   \includegraphics[height=0.13\linewidth,width=0.16\linewidth]{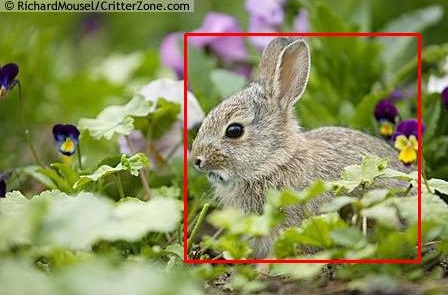}  \ \\
   %
\end{tabular}
\end{center}
\caption{Left to right: Test image and corresponding similar examples.
\label{fig:retrieval_images}}
\end{figure}
\subsection{Internal Saliency Map}

Tang et al.~\cite{tang2014co} present a joint image-box formulation to localize objects from different images.
Inspired by their work, we compute the saliency values of superpixels at each layer by jointly optimizing the above three terms:
%
%
%
\begin{equation}\label{eighth}
    \mathop {\min }\limits_\ell  {\ell ^T}\left\{ {{\rm U} + \alpha \left( {D - W} \right){\rm{ + }}\varepsilon {{\rm I}_n}} \right\}\ell  - {\ell ^T}\beta m .
\end{equation}
To ensure the invertibility of $\left\{ {{\rm U} + \alpha \left( {D - W} \right)} \right\}$, we add a minimum ${\varepsilon {{\rm I}_n}}$ in this quadratic function, where ${{\rm I}_n}\in \mathbb{R}^{n}$ is the identity matrix.
The parameters $\alpha $ and $\beta $ control the tradeoff among these three terms.
Since ${\rm U}$ and ${\left( {D - W} \right)}$ are both positive semi-definite,
the objective function is convex and has a unique solution.
%
%

We compute a saliency map ${{{\tilde S}_k}}$ by summing the superpixels values $\ell$ at each layer,
and then take a weighted linear combination as follows:
\begin{equation}\label{eleventh}
    {S_{-I}} = \frac{1}{6}\sum\nolimits_{k = 1}^6 {\mu _k^s{{\tilde S}_k}} .
\end{equation}
${\mu _k^s}$ controls the weight of different layers, and ${S_{-I}}$ is the final internal saliency map.
\section{Image Retrieval Framework}

The internal saliency map can locate objects with great accuracy by considering the prior, discriminability and similarity information simultaneously. However, a low-level saliency method loses object concepts and may be sensitive to high frequency background noise when the scenes are challenging.
Since similar images with bounding box annotations provide much object information for the test image,
we design an image retrieval framework that searches for similar examples from the validation set of CLS-LOC~\cite{deng2009imagenet} database to further improve the detection performance.
There are 1000 object categories, with 50 validation images for each synset, annotated in the validation set.

The image retrieval framework utilizes pre-stored object regions extracted from similar examples as Linear SVM training samples to learn a linear classifier to predict saliency values of object proposals in the test image, and computes an external saliency map by the sum of regional values. The detailed procedures are listed as follows:

\subsection{Similar Image Retrieval}

For each example image from the dataset, we extract a 4096-dimensional feature vector $x_i^I\in \mathbb{R}^{4096}$ using the pre-trained Caffe framework~\cite{jia2014caffe} and store it.
The similarity of each pair of images is measured by the Euclidean distance between their Caffe features:
\begin{equation}\label{11th}
    {E_{i,j}} = \sum\limits_{k = 1}^{4096} {{{\left| {x_{i,k}^I - x_{j,k}^I} \right|}^2}} .
\end{equation}
%
%
We sort all the examples by their distance to the test image in a descending order and select the top five for subsequent SVM training.
Five similar images provide enough object proposals to train a robust classifier. Furthermore, images with large appearance variations may influence the quality of training samples.
We experimentally find that using five similar images achieves the best performance.
Figure~\ref{fig:retrieval_images} shows some retrieval results.
For most images there are a sufficient number of similar examples, but exceptions do exists such as the third image in the last row.
%
%
\subsection{Region Selection}

\begin{figure}
\begin{center}
\begin{tabular}{@{}c@{}c}
\includegraphics[width=0.49\linewidth]{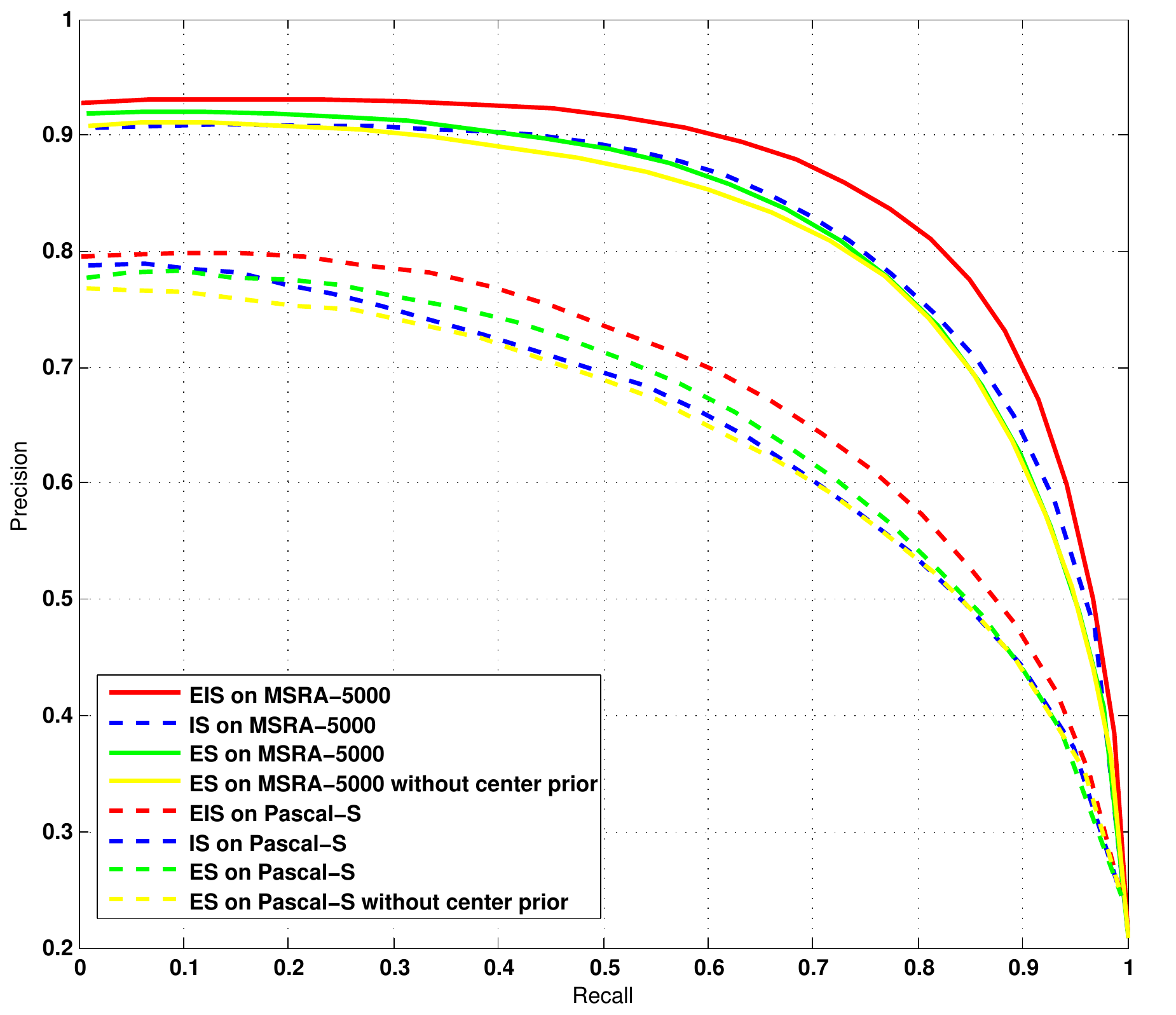} \ &
\includegraphics[width=0.49\linewidth,height=0.42\linewidth]{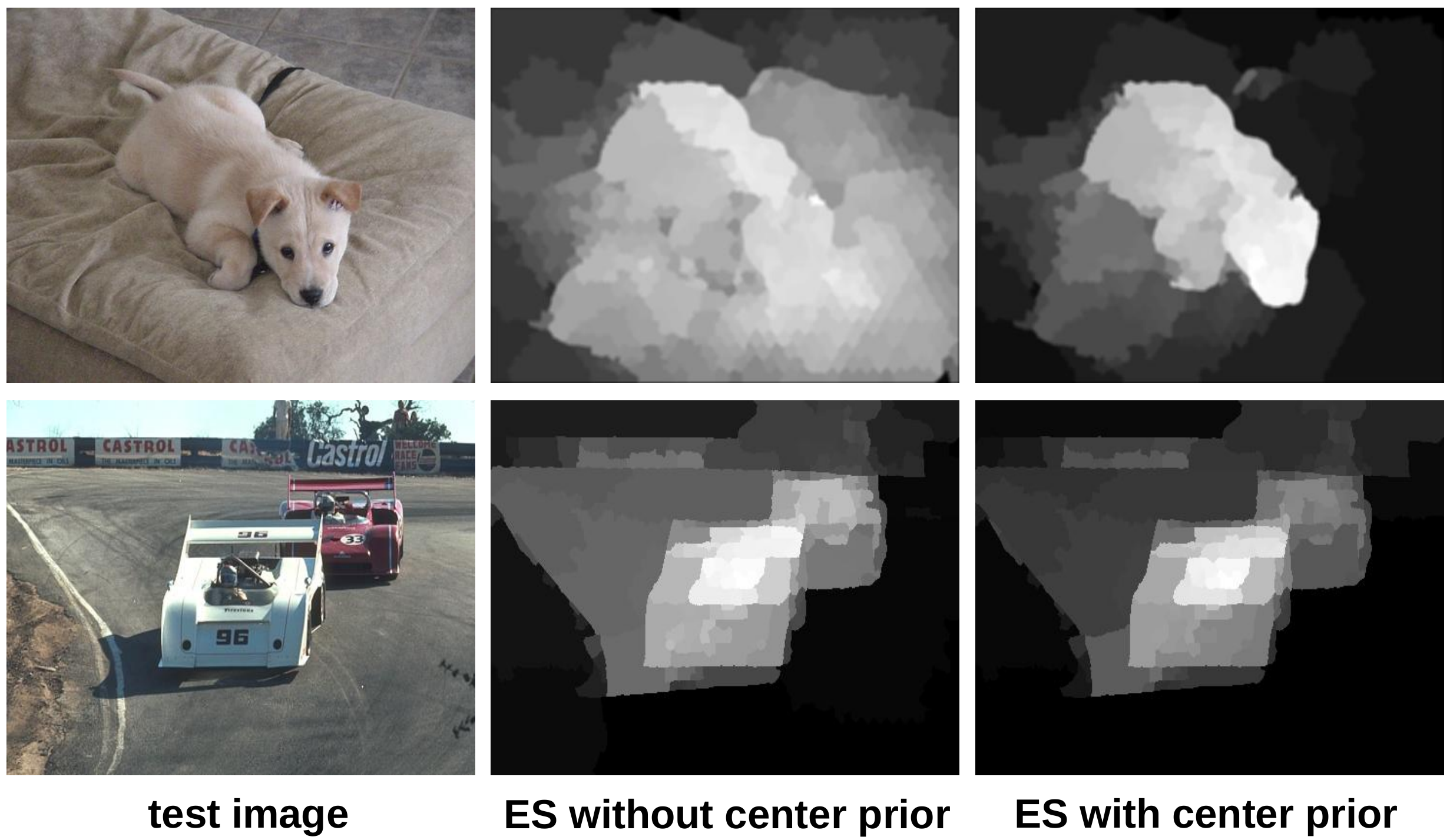}   \ \\
{\small(a)} & {\small(b)}\
\end{tabular}
\end{center}
    \caption{Evaluation of saliency maps: (a) The precision-recall curves of IS, ES (with and without center prior), and EIS on the MSRA-5000 and Pascal-S datasets. (b) Two example results of the ES with and without center prior.
\label{fig:diffmodel}}
\end{figure}

For each test image, we produce a set of object segments using the geodesic object proposal (GOP)~\cite{krahenbuhl2014geodesic} method.
The choice of GOP over other segmentation approaches is motivated by the fact that GOP achieves significantly higher accuracy and runs
substantially faster. For the facility of computation, we select $N$ candidate regions that could potentially contain an object according to their confidence values:
\begin{equation}\label{seventh}
{\eta _r} = \frac{{(1 + \tau ) \times \sum\nolimits_p {\Psi (p) \times {R_r}(p)} }}{{\tau \sum\nolimits_p {\Psi (p)}  + \sum\nolimits_p {{R_r}(p)} }} ,
\end{equation}
where
\begin{equation}\label{sixth}
\Psi  = {S_{-I}} \times {\cal O} \times G .
\end{equation}
${\cal O}$ is the objectness map and ${{R_r}}$ is the mask of region $r$.
${R_r}(p) = 1$ indicates that pixel $p$ belongs to $r$, and ${R_r}(p) = 0$ otherwise.
\begin{equation}\label{sixth2}
    G = \exp \left\{ { - \frac{{{{\left( {{x_p} - {x_c}} \right)}^2}}}{{2{\sigma _x}^2}} - \frac{{{{\left( {{y_p} - {y_c}} \right)}^2}}}{{2{\sigma _y}^2}}} \right\}
\end{equation}
is a center prior map, where ${{x_p}}$ and ${{y_p}}$ are the coordinates of pixel $p$, ${{x_c}}$ and ${{y_c}}$ denote the center of test image, and ${{\sigma _x}}$ and ${{\sigma _y}}$ are weight parameters to control the strength of distances.
Two example results and PR curves in Figure~\ref{fig:diffmodel} demonstrate the efficiency of the center prior map.
We experientially set $N=100$ and $\tau  = 0.4$ in all experiments. The external saliency map is constructed based on these selected region proposals.
\subsection{Regional Features}

Different features affect the performance of vision tasks significantly. Therefore, designing discriminative features is essential to our work.
In this part, we propose a 81-dimensional feature vector, $x_i^r\in \mathbb{R}^{81}$, to describe each region.
The detailed components of regional features are listed in Table~\ref{fig:table1}.
We define the 15-pixel wide narrow border regions of the test image as background regions.
The color histogram and mean color distances are measured by the chi-square and Euclidean distances between each candidate proposal and the background regions respectively.
We also add the superpixel features, replacing superpixels with regions, and 3-dimensional shape features to the component list to form a 81-dimensional feature vector.
Regions extracted from the test image or its similar examples are represented by this feature vector.
%
\begin{figure*}
\begin{center}
\begin{tabular}{@{}c@{}c@{}c@{}c@{}c@{}c}
\includegraphics[width=0.25\linewidth]{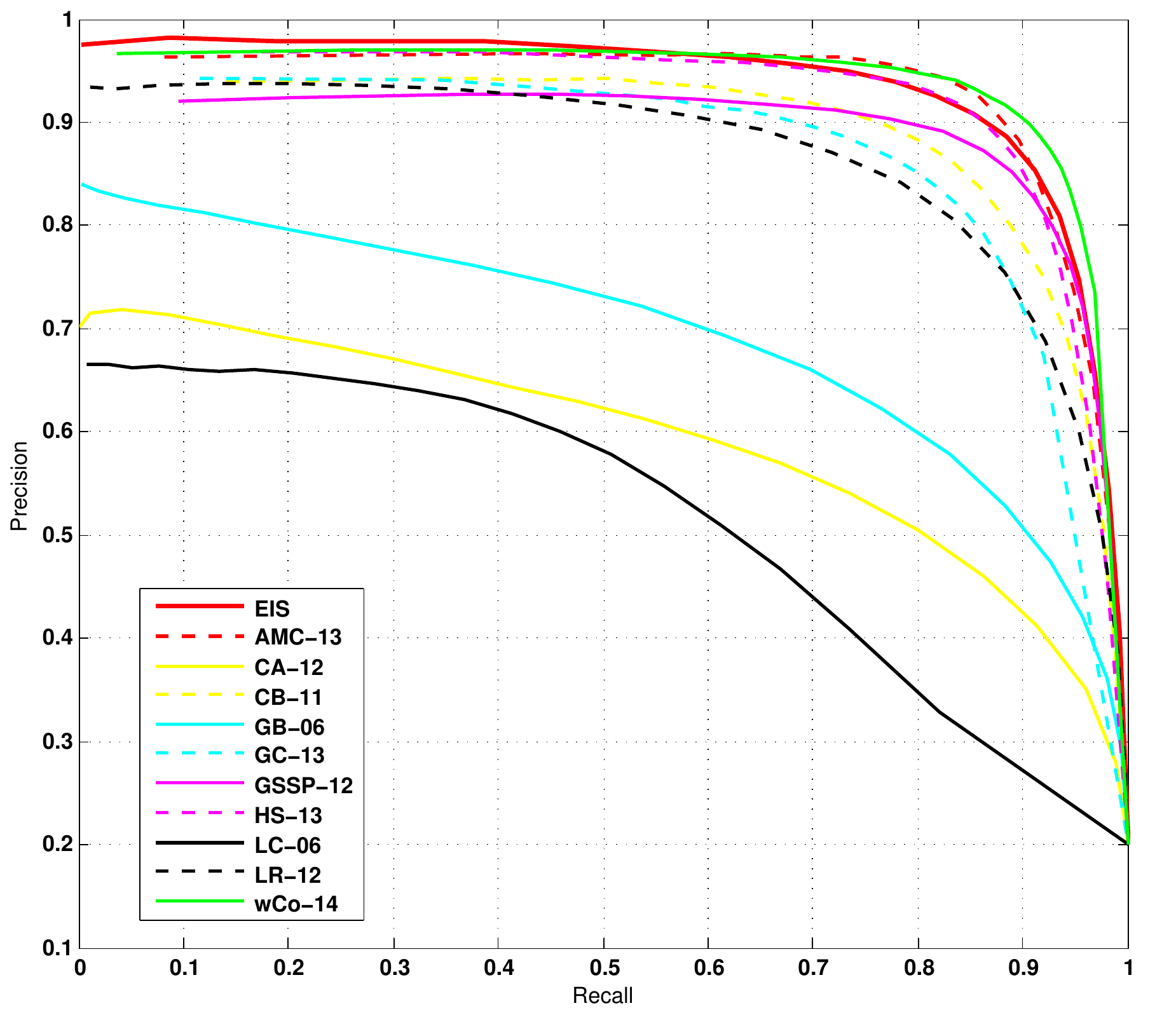} \ &
\includegraphics[width=0.25\linewidth]{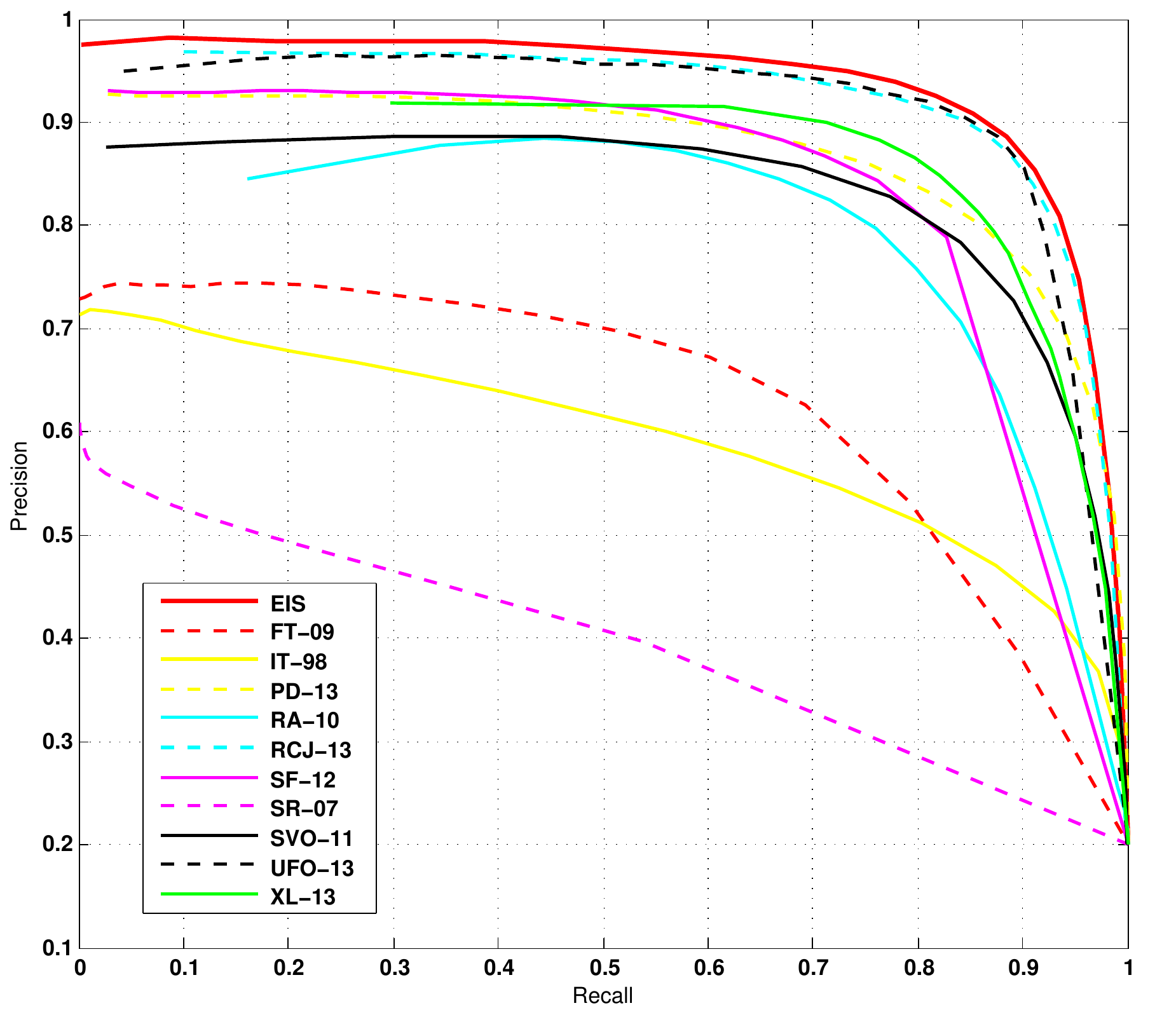}   \ &
\includegraphics[width=0.25\linewidth]{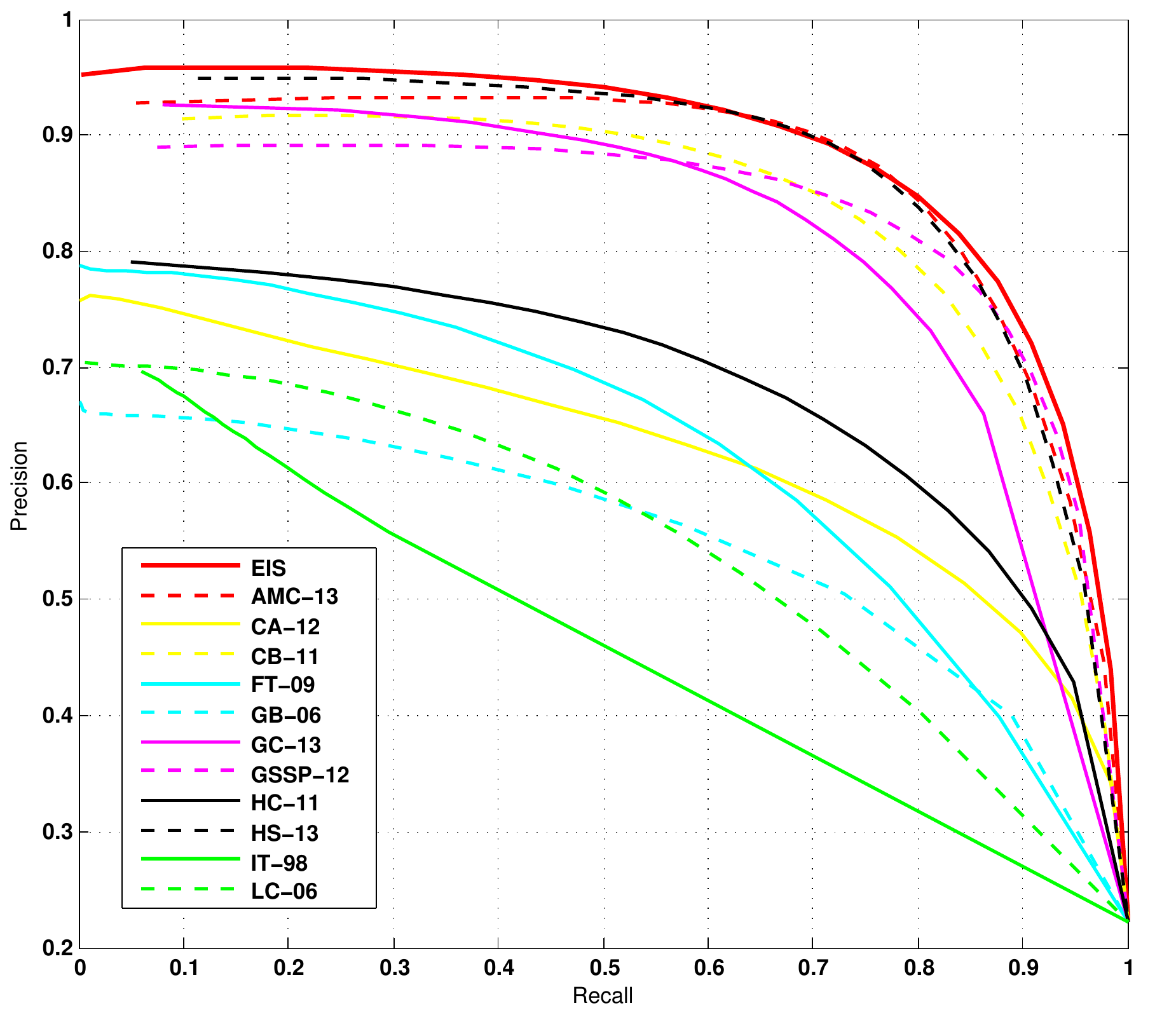}   \ &
\includegraphics[width=0.25\linewidth]{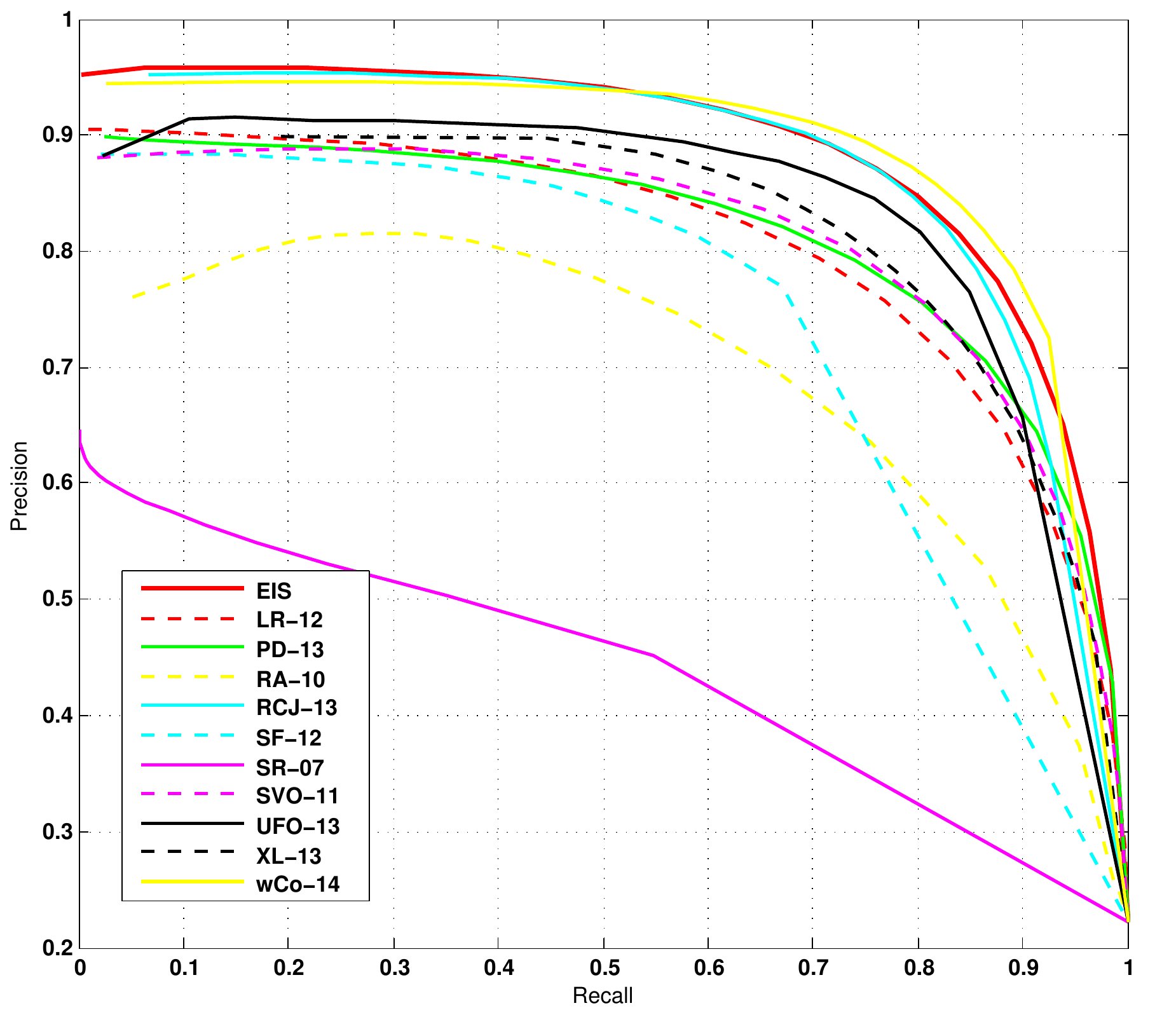}   \ \\
{\small(a)} & {\small(b)} & {\small(c)} & {\small(d)}\
\end{tabular}
\end{center}
    \caption{Results of different methods: (a), (b) Precision-recall curve on the ASD dataset. (c), (d) Precision-recall curves on the THUS dataset.
\label{fig:pr1}}
\end{figure*}
\begin{figure*}
\begin{center}
\begin{tabular}{@{}c@{}c@{}c@{}c@{}c@{}c}
\includegraphics[width=0.25\linewidth]{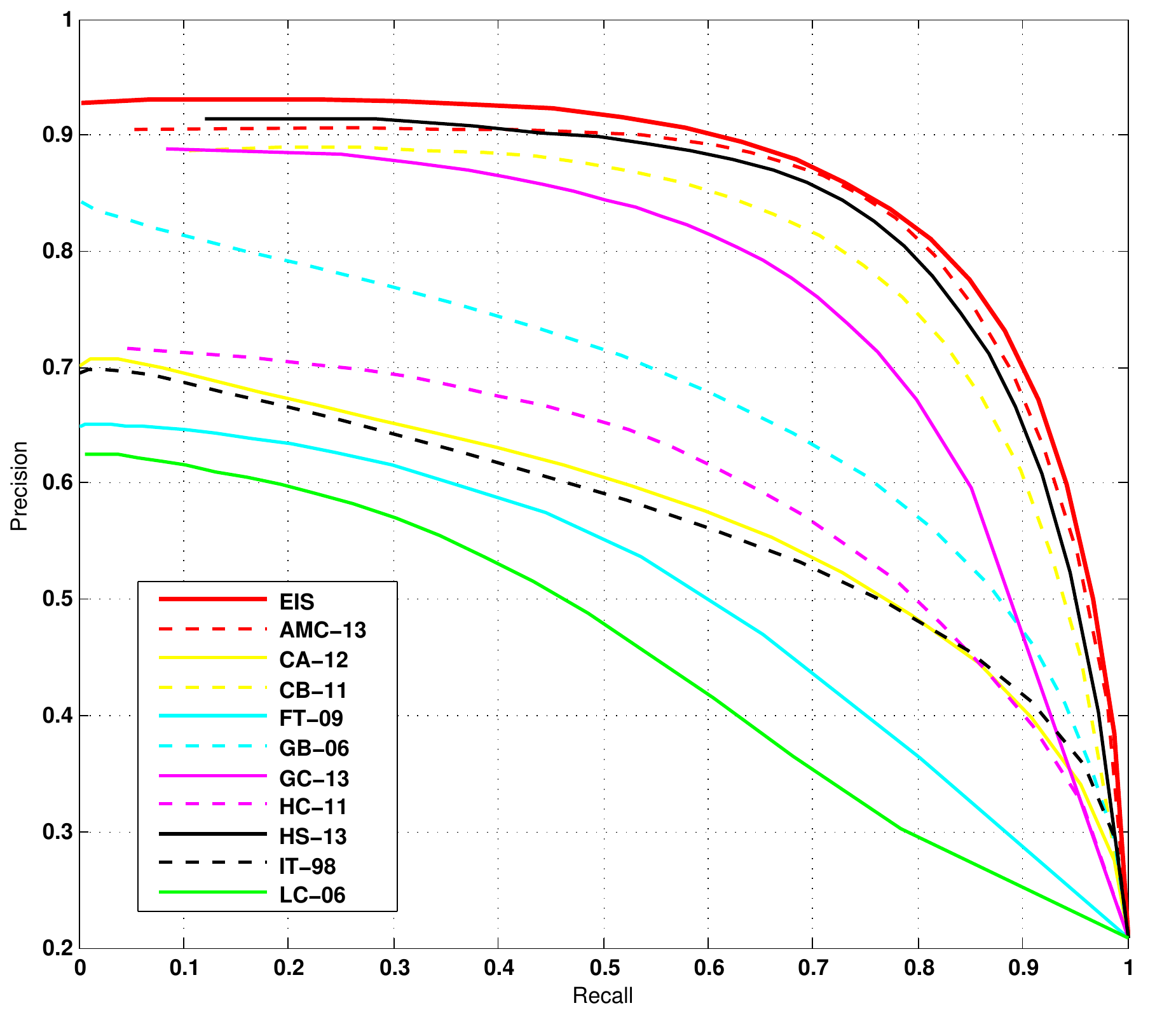} \ &
\includegraphics[width=0.25\linewidth]{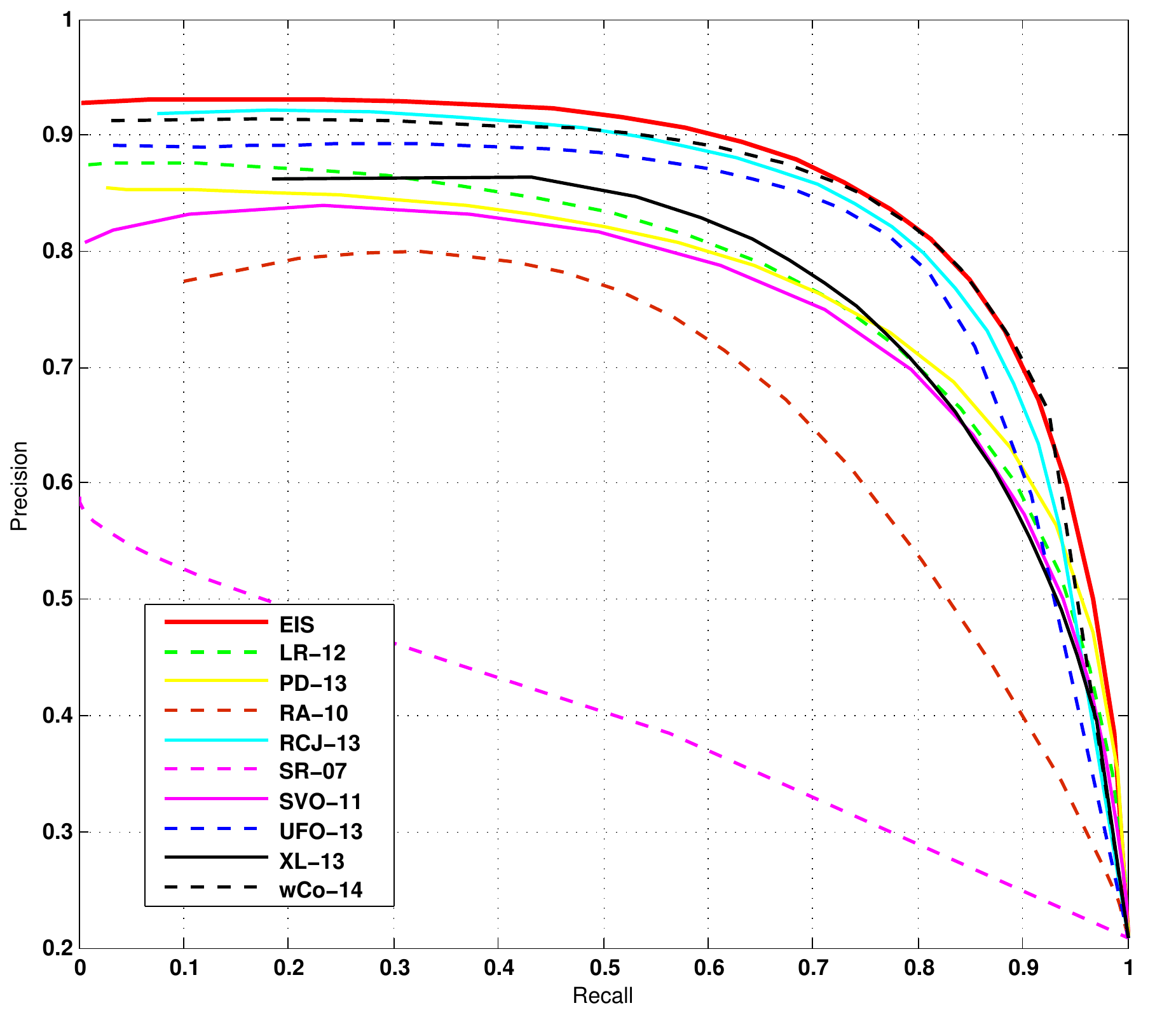}   \ &
\includegraphics[width=0.25\linewidth]{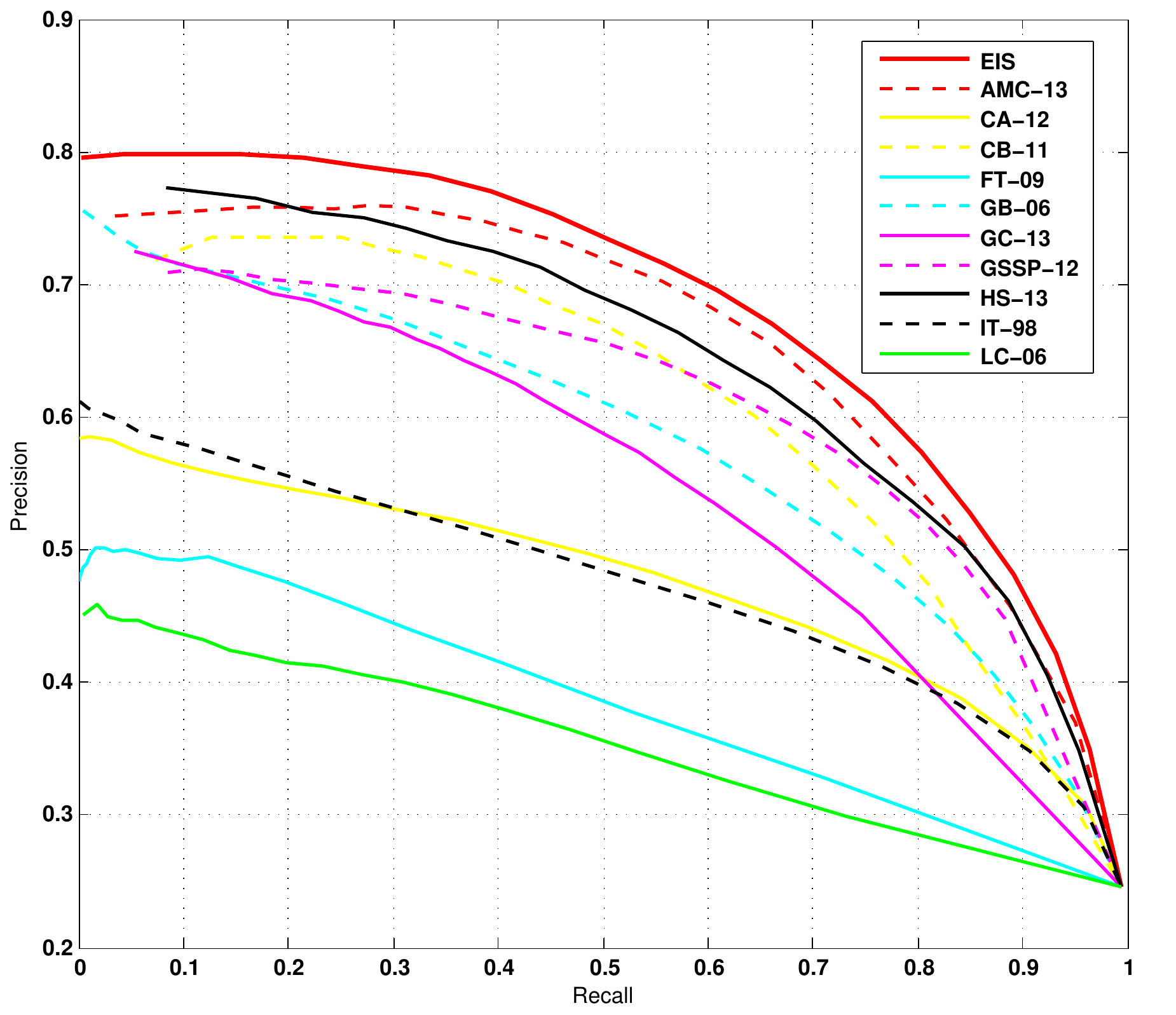}   \ &
\includegraphics[width=0.25\linewidth]{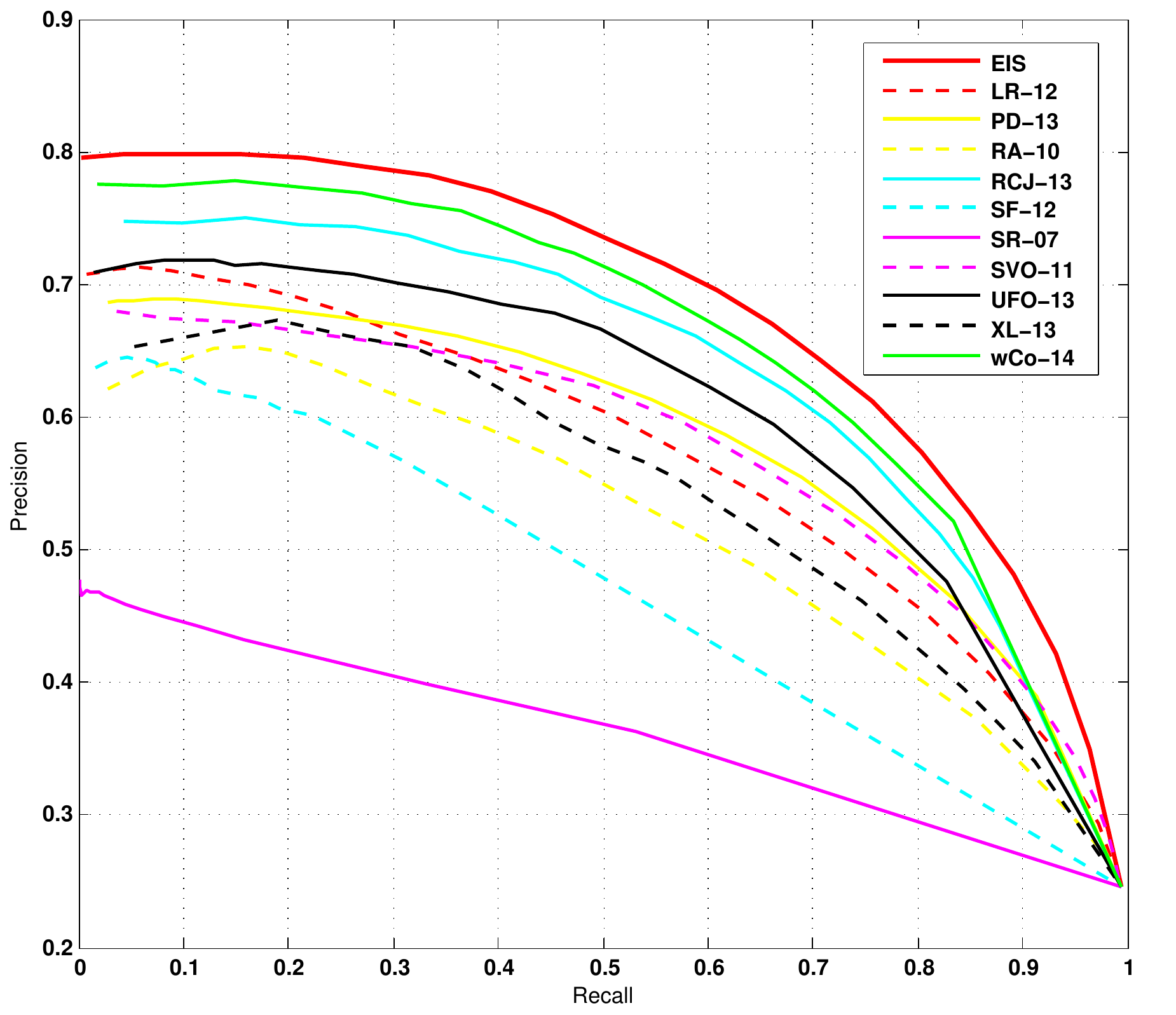}   \ \\
{\small(a)} & {\small(b)} & {\small(c)} & {\small(d)}\
\end{tabular}
\end{center}
    \caption{Results of different methods: (a), (b) Precision-recall curve on the MSRA-5000 dataset. (c), (d) Precision-recall curves on the Pascal-S dataset.
\label{fig:pr2}}
\end{figure*}
\begin{figure*}
\begin{center}
\begin{tabular}{@{}c@{}c@{}c@{}c@{}c@{}c}
\includegraphics[width=0.25\linewidth]{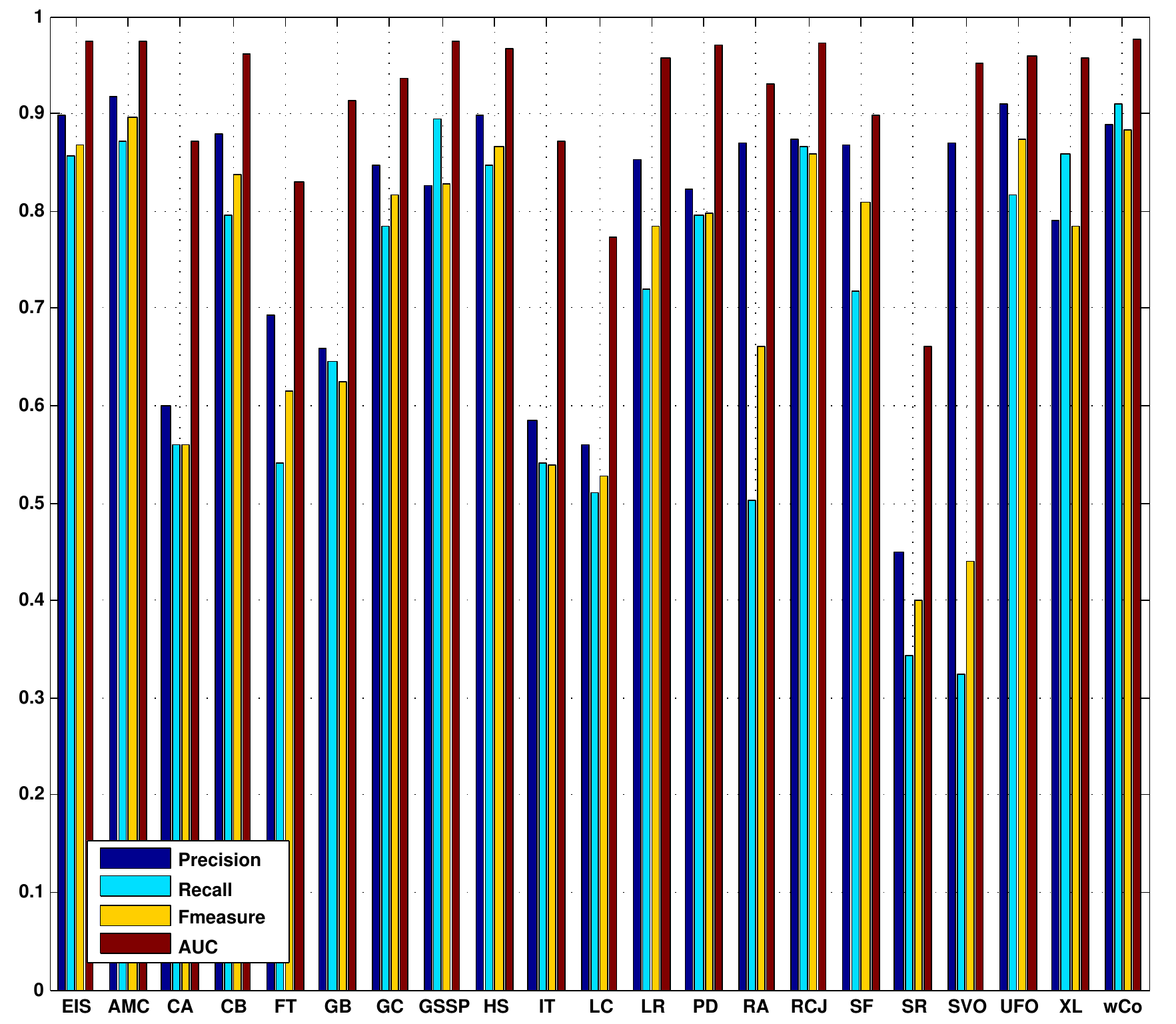} \ &
\includegraphics[width=0.25\linewidth]{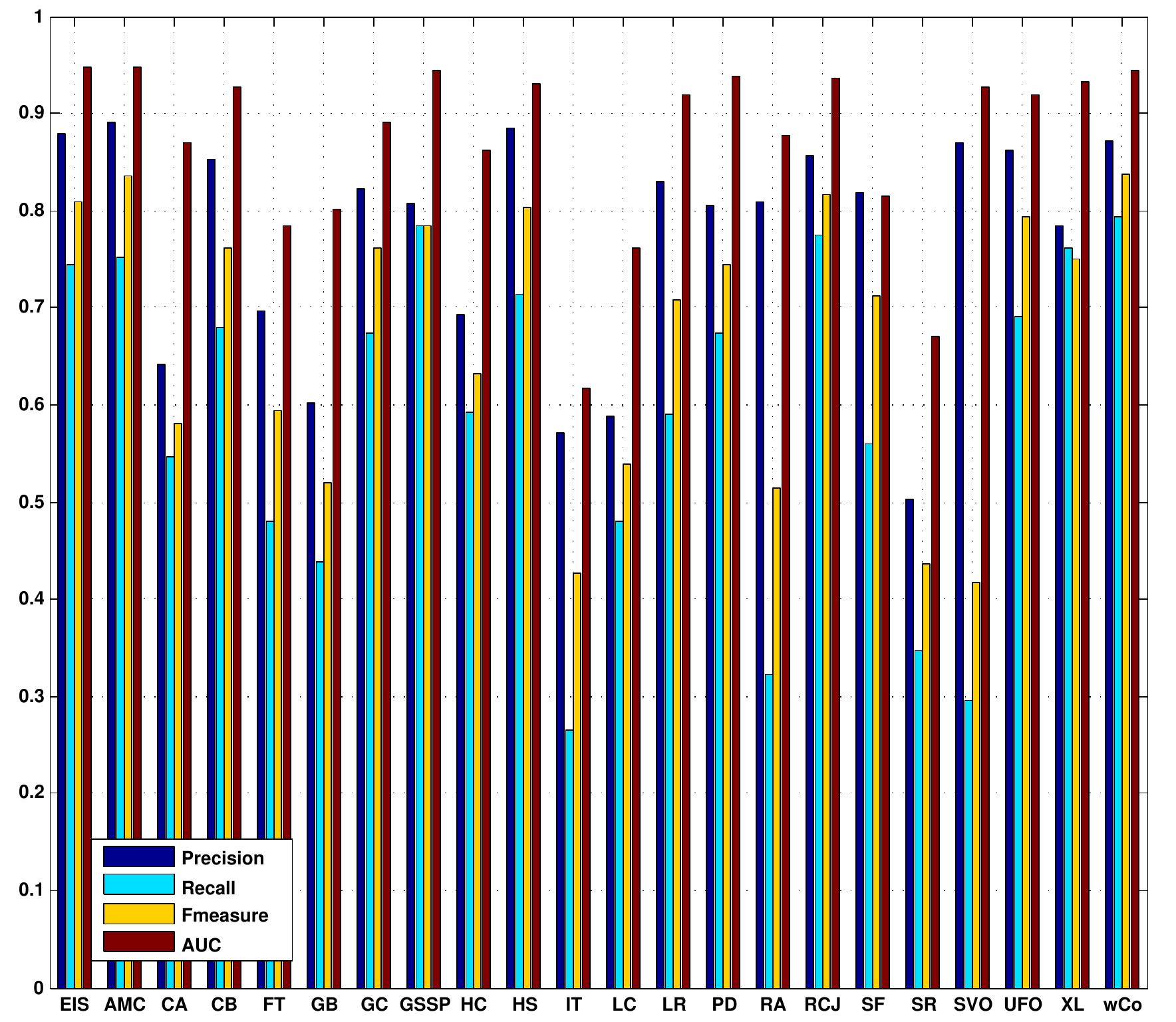} \ &
\includegraphics[width=0.25\linewidth]{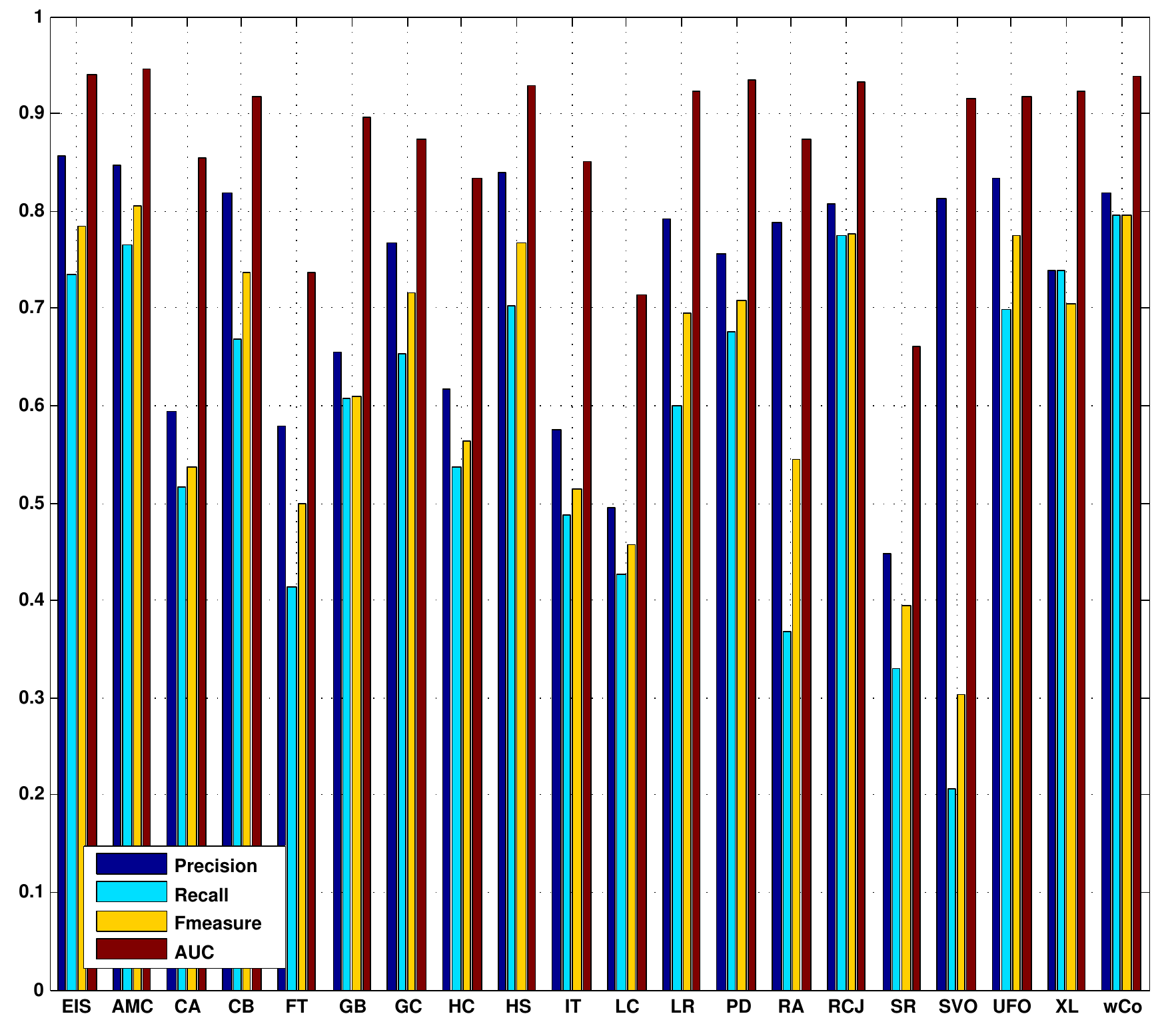} \ &
\includegraphics[width=0.25\linewidth]{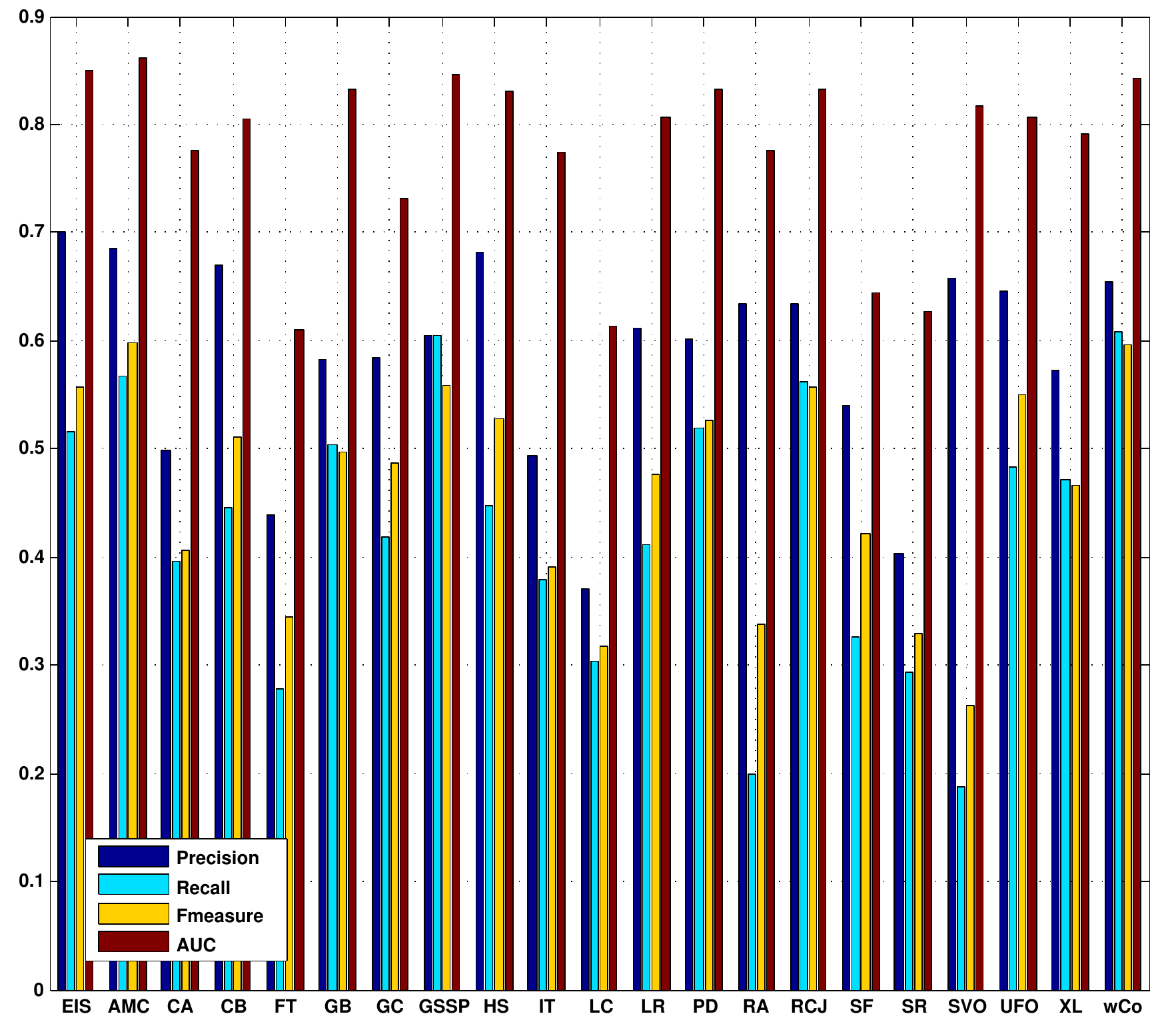} \ \\
{\small ASD} & {\small THUS} & {\small MSRA-5000} & {\small Pascal-S}\
\end{tabular}
\end{center}
    \caption{Average precision, recall, F-Measure and AUC of methods on different datasets.
\label{fig:bar}}
\end{figure*}

\subsection{External Saliency Map}

Instead of utilizing the whole dataset as training samples, we select a subset of similar images to train a customized SVM for each test image.
Images from the CLS-LOC database are segmented into object proposals by the GOP, with each proposal corresponding to a saliency label $\zeta $ based on its overlapping area with the ground truth bounding boxes.
To save time, we pre-store these segments, with labels and regional features, and load them directly once the corresponding image is selected as one of similar examples.

We learn parameters, $\nu $ and $b$, by training a linear classifier $f(x) = {\nu ^T}x + b$, and predict the saliency value of each candidate region in the test image as follows:
\begin{equation}\label{twenty}
    {\zeta _k} = {\nu ^T}x_k^r + b .
\end{equation}
The external saliency map is generated by adding the regional values of 100 selected proposals:
\begin{equation}\label{twentyone}
    {S_{-E}} = \sum\nolimits_{k = 1}^{100} {{\zeta _k}{\mu _k^r}} ,
\end{equation}
where ${\mu _k^r}$, having the same size with the test image, is the mask map of region $k$.
The external saliency map can locate salient objects accurately in most cases, which demonstrates the efficiency of the proposed image retrieval framework.
\section{Final Saliency Map (EIS)}

The image retrieval framework adopts a supervised learning approach to address saliency detection, which contains the high-level object concept and achieves good performance in localizing salient objects.
The external saliency map can uniformly highlight the whole salient region with explicit object boundaries except in the case when the test
image cannot find similar examples.
As an essential supplement to image retrieval, an internal optimization module is proposed and combined with the external saliency map to construct the final saliency map.
The internal saliency map captures low-level feature contrast within an image and performs well in identifying the salient superpixels.
But it easily gets affected by background noise, especially when dealing with challenging scenes.
To make the best use of their advantages, we propose to take a weighted sum of the internal and external saliency maps:
%
%
\begin{equation}\label{twentytwo}
    S = \gamma {S_{-E}} + (1 - \gamma ){S_{-I}} ,
\end{equation}
where $\gamma $ controls the tradeoff between these two maps, and $S$ is the final saliency map of our method.
\begin{figure*}[t]
\begin{center}
\begin{tabular}{@{}c@{}c@{}c@{}c@{}c@{}c@{}c@{}c@{}c@{}c@{}c}
   \includegraphics[width=0.0865\linewidth]{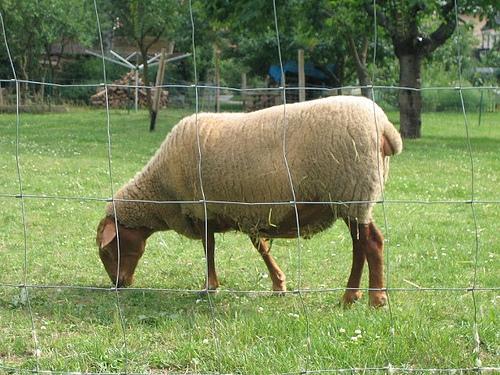}  \ &
   \includegraphics[width=0.0865\linewidth]{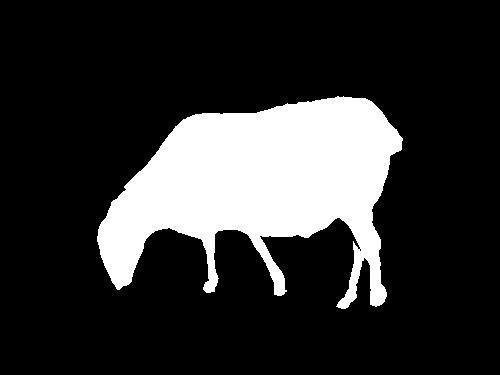}  \ &
   \includegraphics[width=0.0865\linewidth]{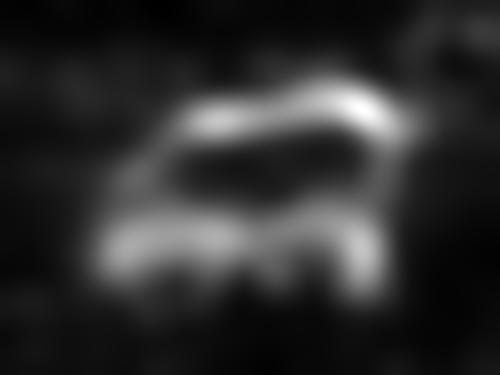}  \ &
   \includegraphics[width=0.0865\linewidth]{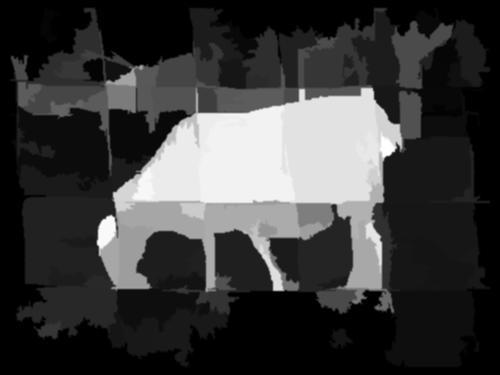}  \ &
   \includegraphics[width=0.0865\linewidth]{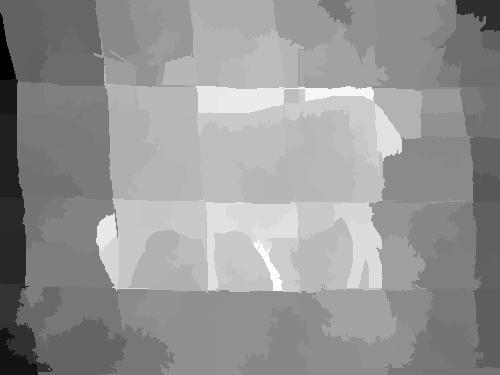}  \ &
   \includegraphics[width=0.0865\linewidth]{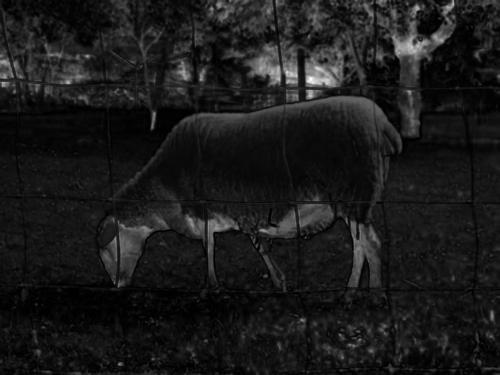}  \ &
   \includegraphics[width=0.0865\linewidth]{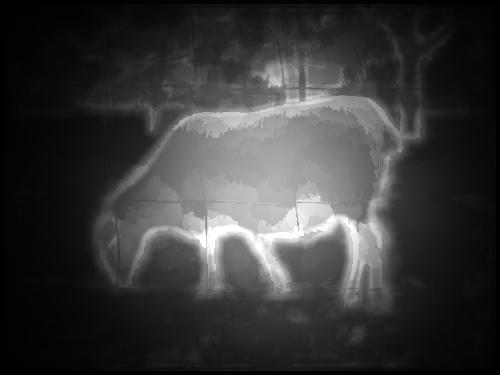}  \ &
   \includegraphics[width=0.0865\linewidth]{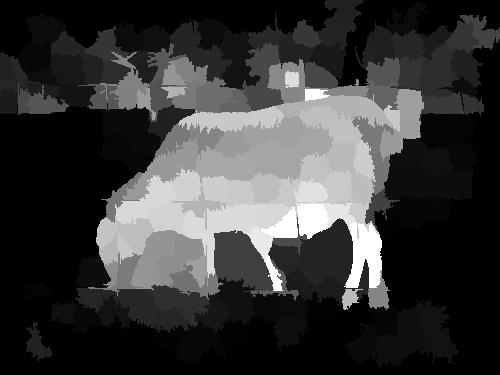}  \ &
   \includegraphics[width=0.0865\linewidth]{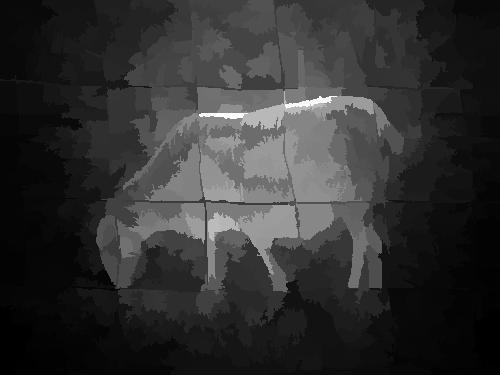}  \ &
   \includegraphics[width=0.0865\linewidth]{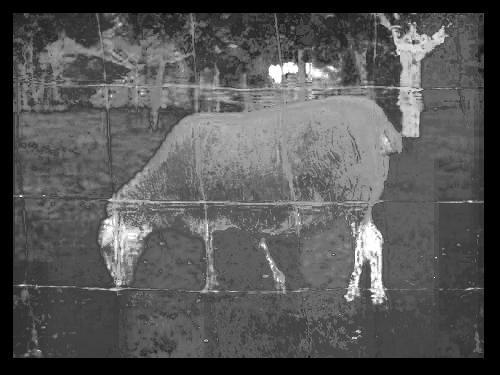}  \ &
   \includegraphics[width=0.0865\linewidth]{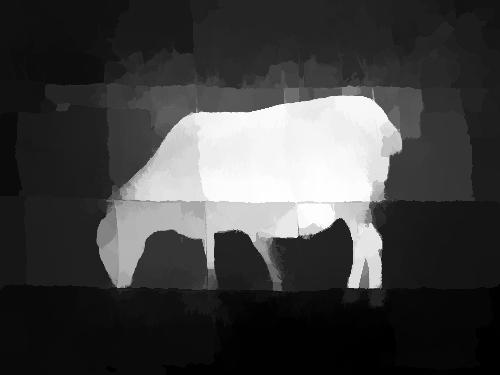}  \ \\
   \includegraphics[width=0.0865\linewidth]{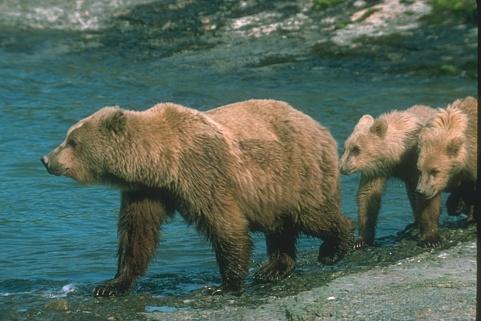}  \ &
   \includegraphics[width=0.0865\linewidth]{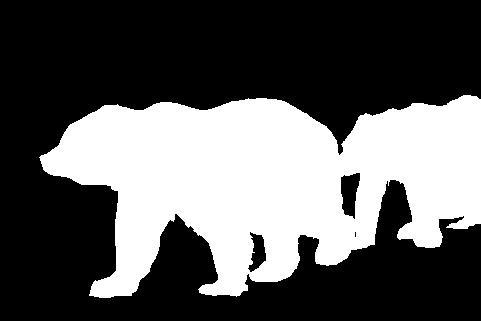}  \ &
   \includegraphics[width=0.0865\linewidth]{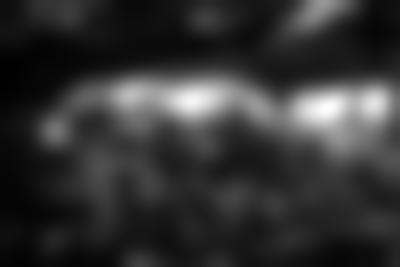}  \ &
   \includegraphics[width=0.0865\linewidth]{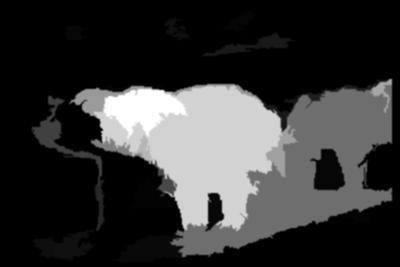}  \ &
   \includegraphics[width=0.0865\linewidth]{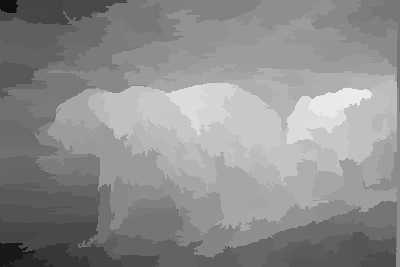}  \ &
   \includegraphics[width=0.0865\linewidth]{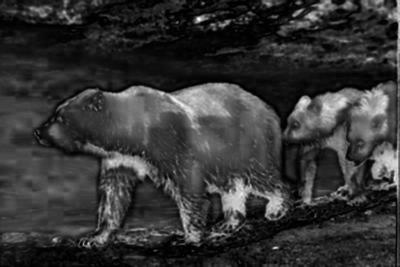}  \ &
   \includegraphics[width=0.0865\linewidth]{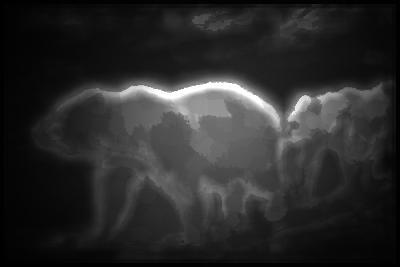}  \ &
   \includegraphics[width=0.0865\linewidth]{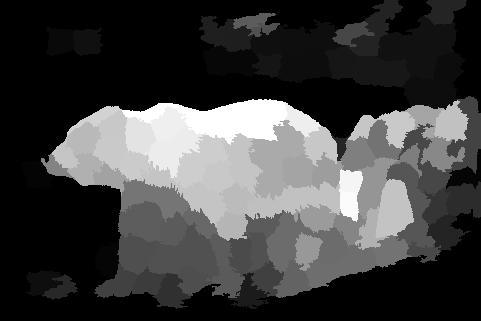}  \ &
   \includegraphics[width=0.0865\linewidth]{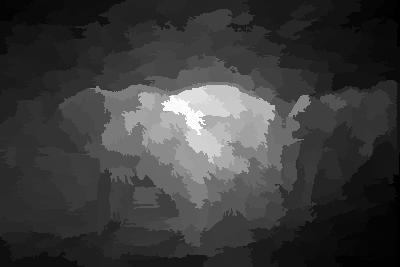}  \ &
   \includegraphics[width=0.0865\linewidth]{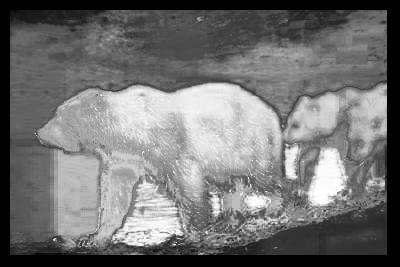}  \ &
   \includegraphics[width=0.0865\linewidth]{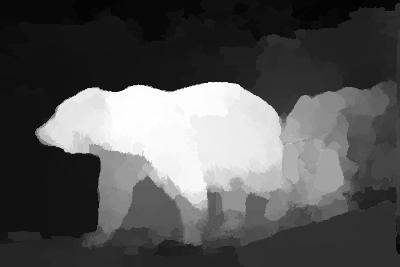}  \ \\
      \includegraphics[width=0.0865\linewidth]{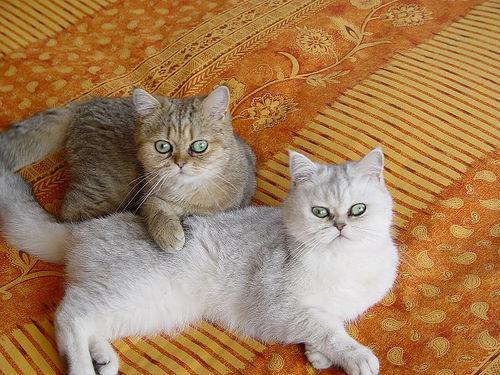}  \ &
   \includegraphics[width=0.0865\linewidth]{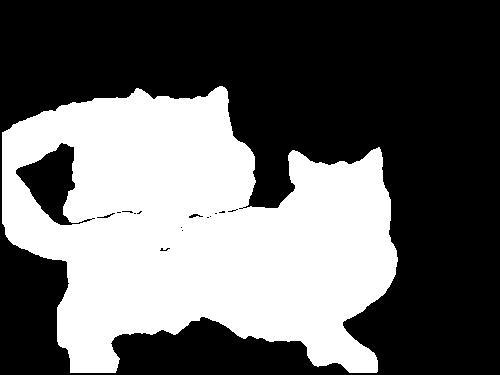}  \ &
   \includegraphics[width=0.0865\linewidth]{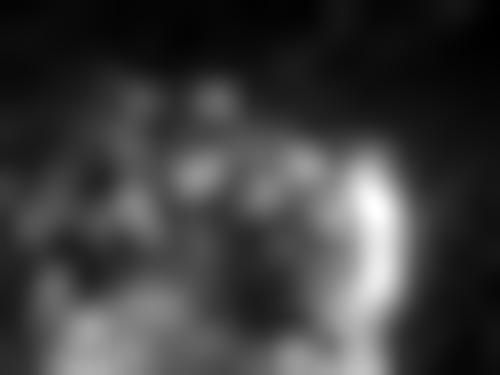}  \ &
   \includegraphics[width=0.0865\linewidth]{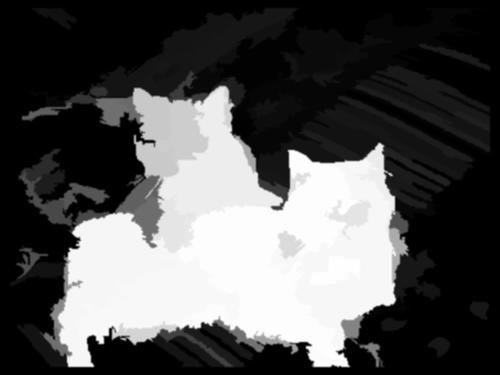}  \ &
   \includegraphics[width=0.0865\linewidth]{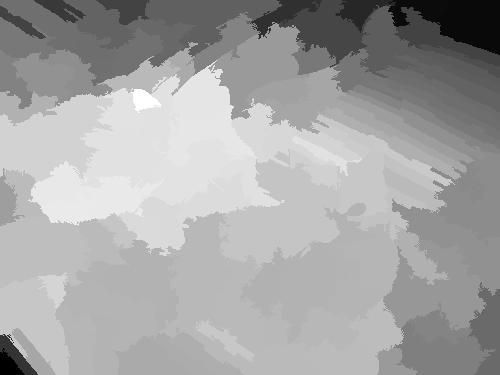}  \ &
   \includegraphics[width=0.0865\linewidth]{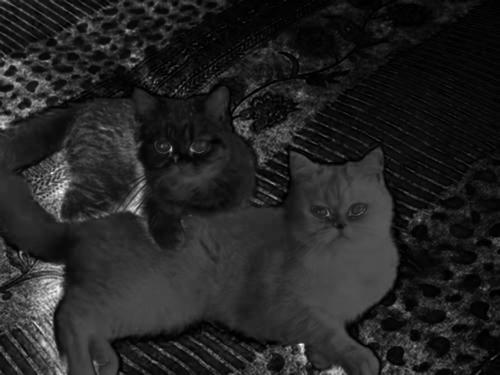}  \ &
   \includegraphics[width=0.0865\linewidth]{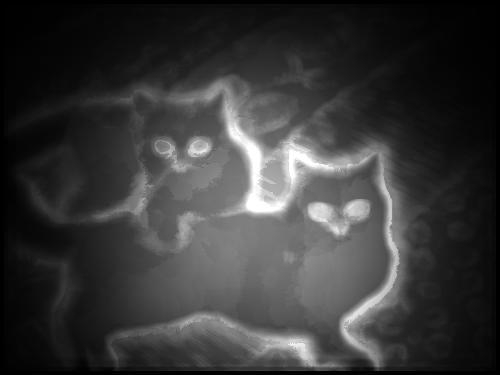}  \ &
   \includegraphics[width=0.0865\linewidth]{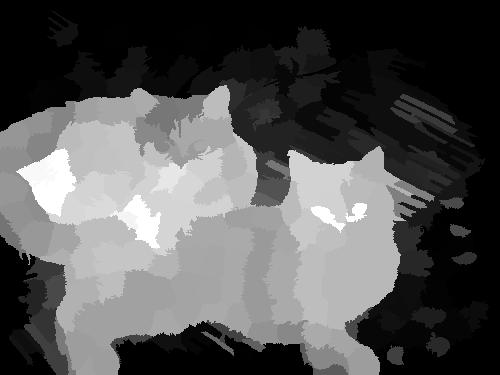}  \ &
   \includegraphics[width=0.0865\linewidth]{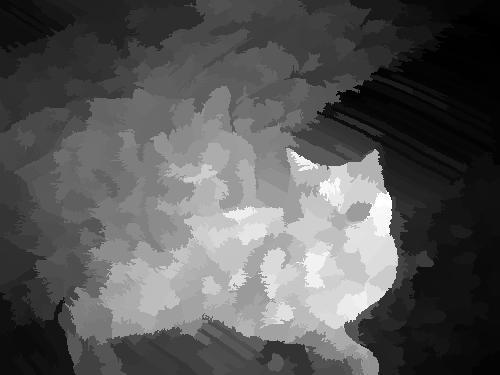}  \ &
   \includegraphics[width=0.0865\linewidth]{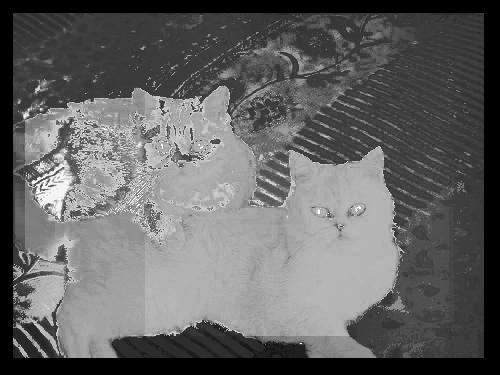}  \ &
   \includegraphics[width=0.0865\linewidth]{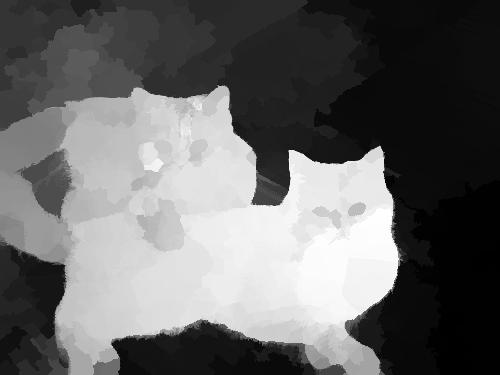}  \ \\
   \includegraphics[width=0.0865\linewidth]{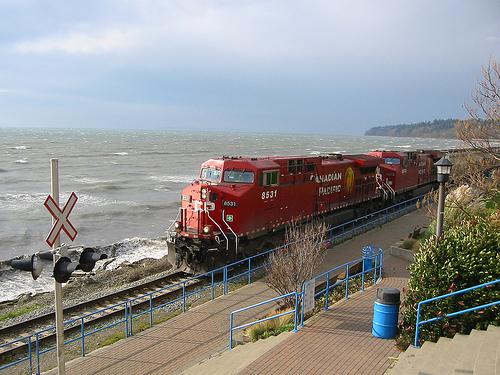}  \ &
   \includegraphics[width=0.0865\linewidth]{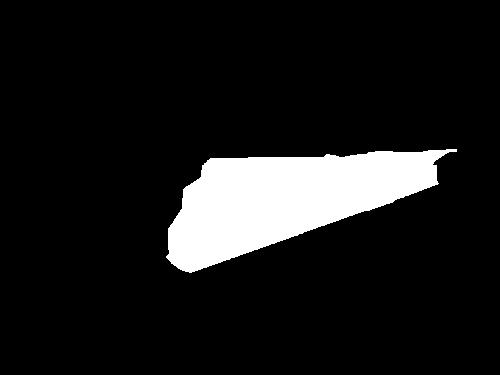}  \ &
   \includegraphics[width=0.0865\linewidth]{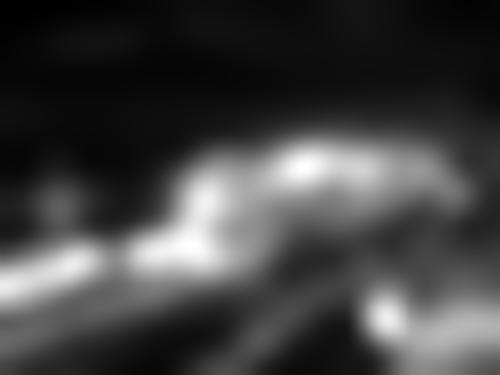}  \ &
   \includegraphics[width=0.0865\linewidth]{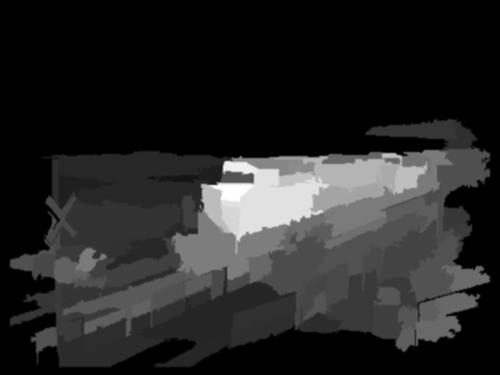}  \ &
   \includegraphics[width=0.0865\linewidth]{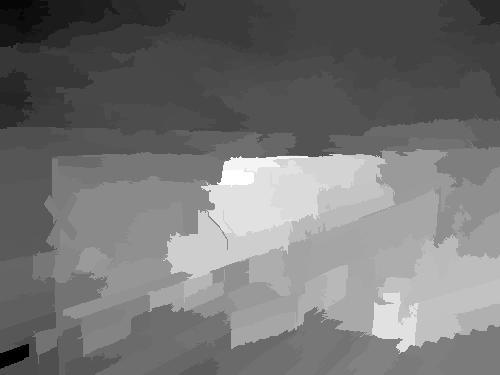}  \ &
   \includegraphics[width=0.0865\linewidth]{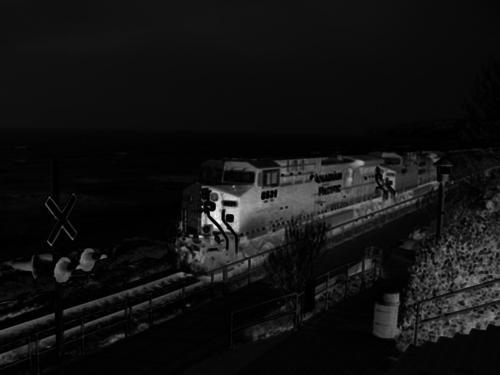}  \ &
   \includegraphics[width=0.0865\linewidth]{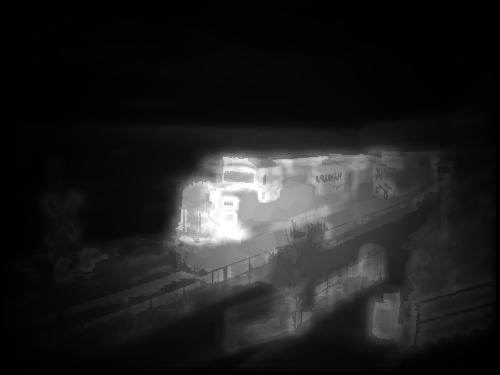}  \ &
   \includegraphics[width=0.0865\linewidth]{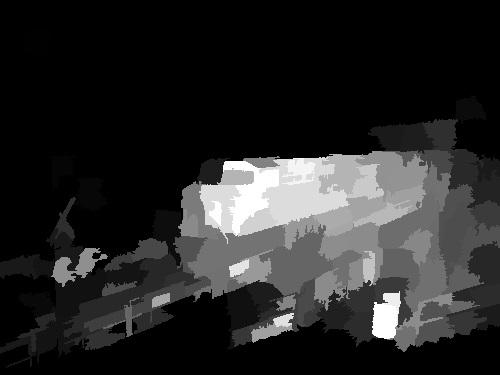}  \ &
   \includegraphics[width=0.0865\linewidth]{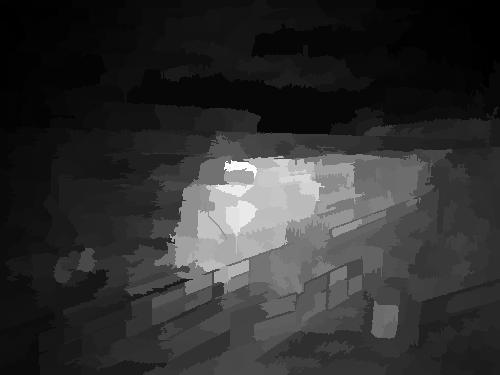}  \ &
   \includegraphics[width=0.0865\linewidth]{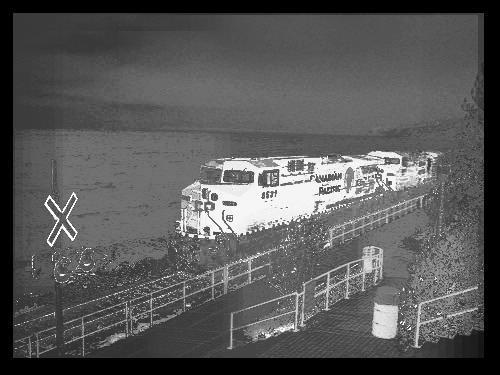}  \ &
   \includegraphics[width=0.0865\linewidth]{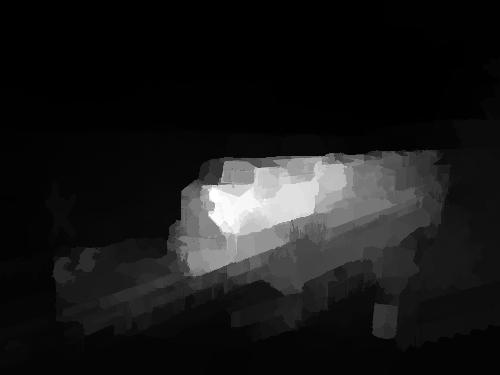}  \ \\
   %
   \includegraphics[width=0.0865\linewidth]{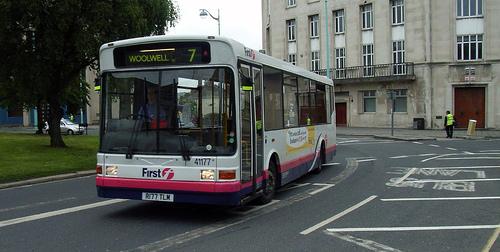}  \ &
   \includegraphics[width=0.0865\linewidth]{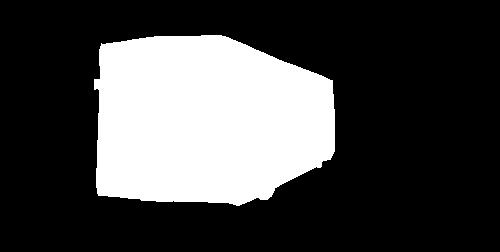}  \ &
   \includegraphics[width=0.0865\linewidth]{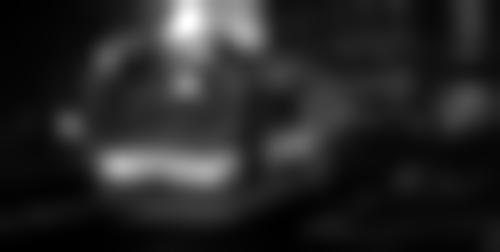}  \ &
   \includegraphics[width=0.0865\linewidth]{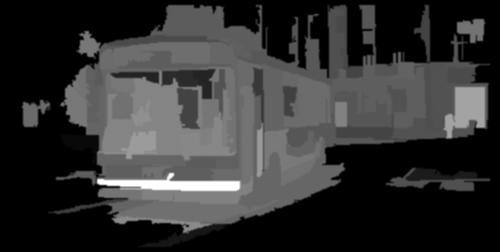}  \ &
   \includegraphics[width=0.0865\linewidth]{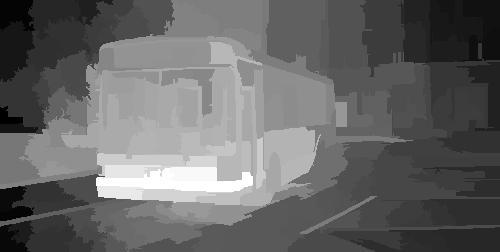}  \ &
   \includegraphics[width=0.0865\linewidth]{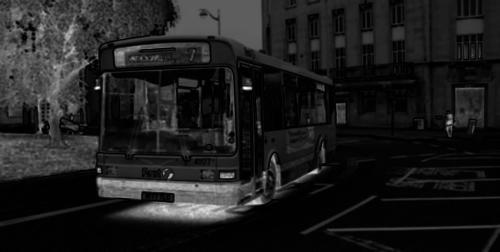}  \ &
   \includegraphics[width=0.0865\linewidth]{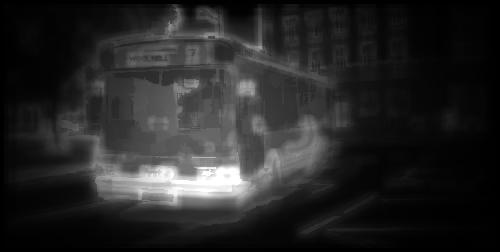}  \ &
   \includegraphics[width=0.0865\linewidth]{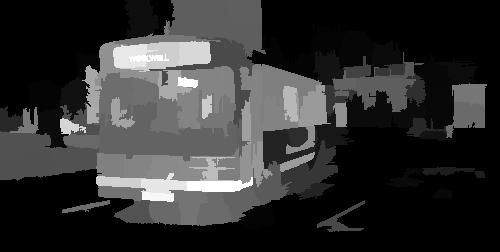}  \ &
   \includegraphics[width=0.0865\linewidth]{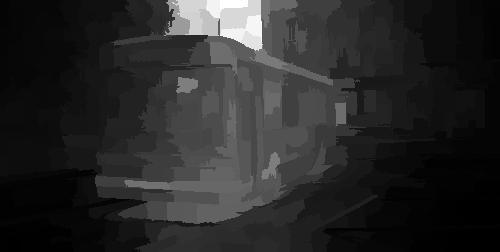}  \ &
   \includegraphics[width=0.0865\linewidth]{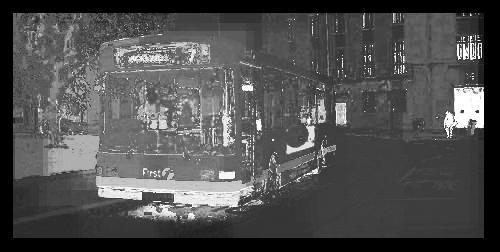}  \ &
   \includegraphics[width=0.0865\linewidth]{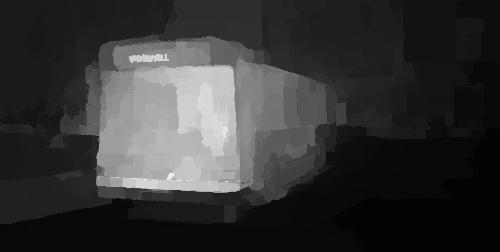}  \ \\
   \includegraphics[width=0.0865\linewidth]{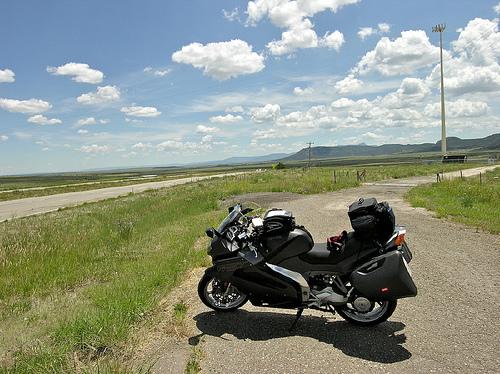}  \ &
   \includegraphics[width=0.0865\linewidth]{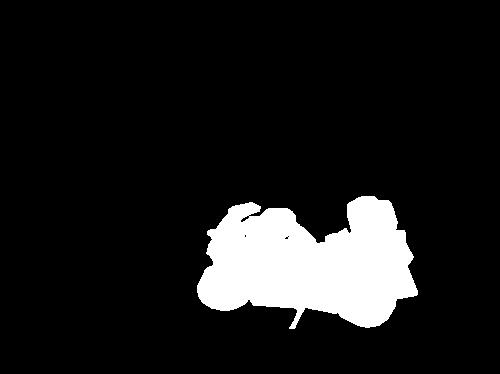}  \ &
   \includegraphics[width=0.0865\linewidth]{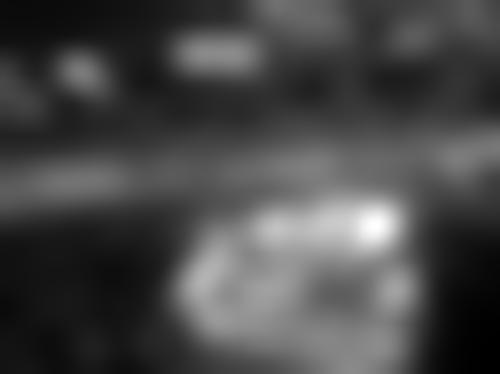}  \ &
   \includegraphics[width=0.0865\linewidth]{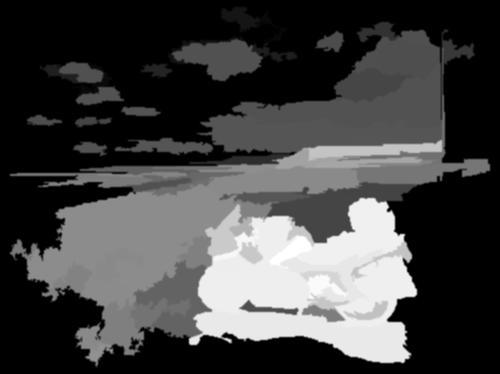}  \ &
   \includegraphics[width=0.0865\linewidth]{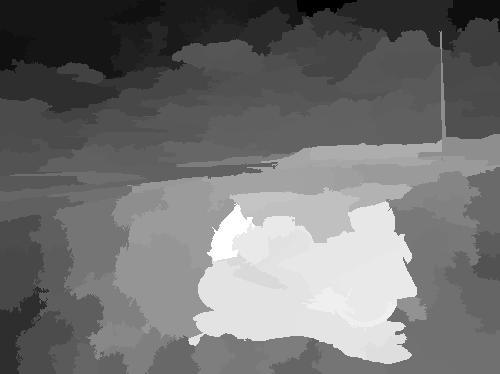}  \ &
   \includegraphics[width=0.0865\linewidth]{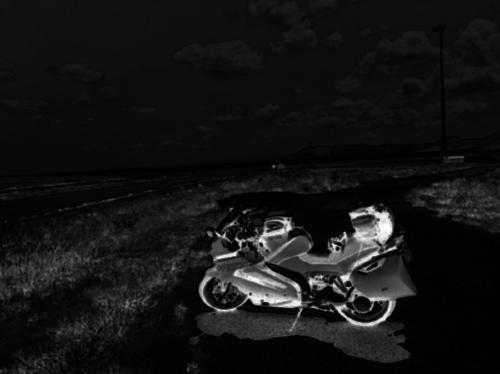}  \ &
   \includegraphics[width=0.0865\linewidth]{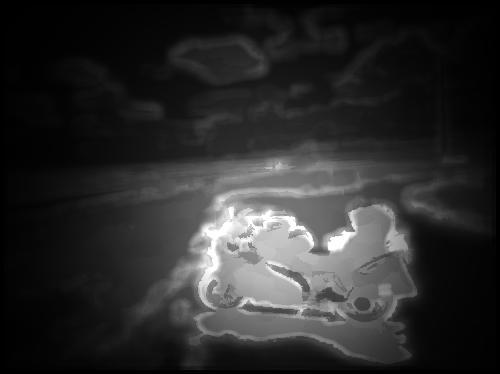}  \ &
   \includegraphics[width=0.0865\linewidth]{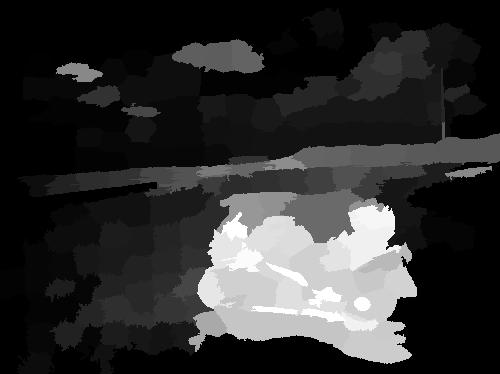}  \ &
   \includegraphics[width=0.0865\linewidth]{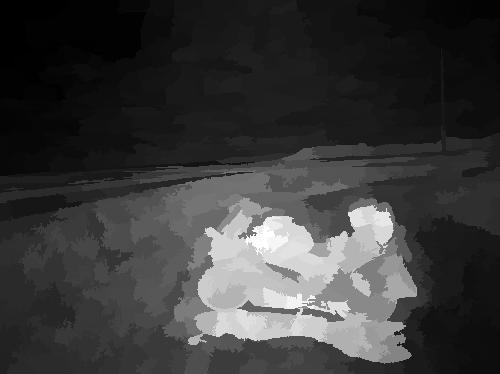}  \ &
   \includegraphics[width=0.0865\linewidth]{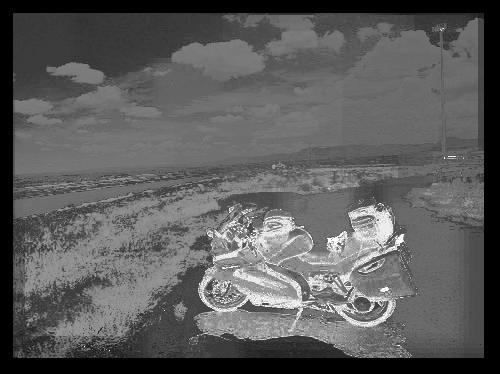}  \ &
   \includegraphics[width=0.0865\linewidth]{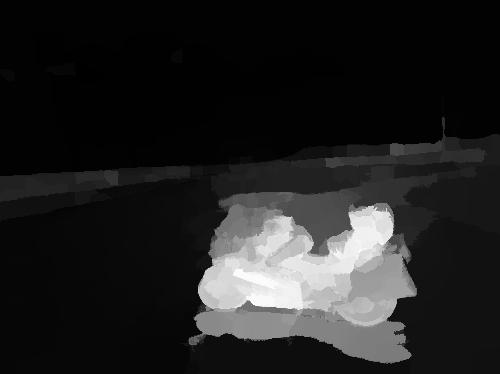}  \ \\
%
   \includegraphics[width=0.0865\linewidth]{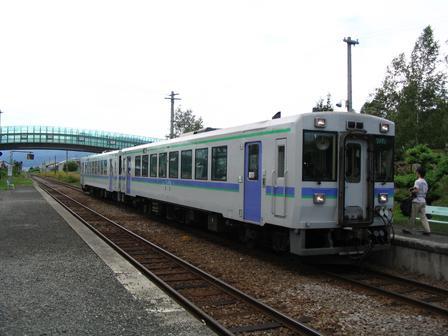}  \ &
   \includegraphics[width=0.0865\linewidth]{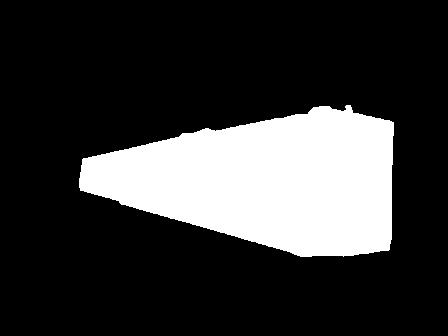}  \ &
   \includegraphics[width=0.0865\linewidth]{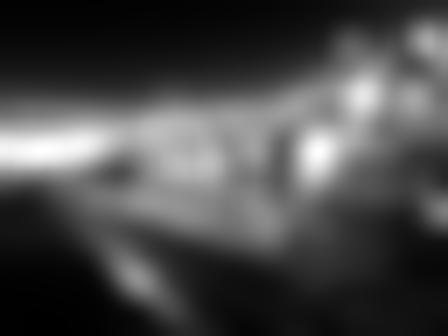}  \ &
   \includegraphics[width=0.0865\linewidth]{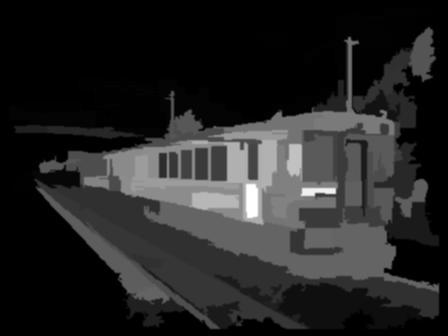}  \ &
   \includegraphics[width=0.0865\linewidth]{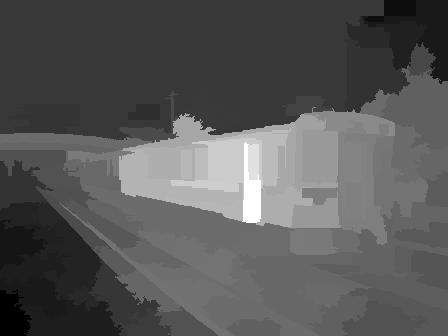}  \ &
   \includegraphics[width=0.0865\linewidth]{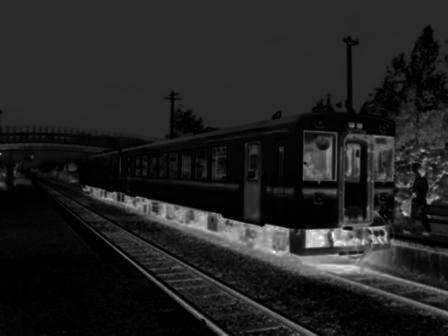}  \ &
   \includegraphics[width=0.0865\linewidth]{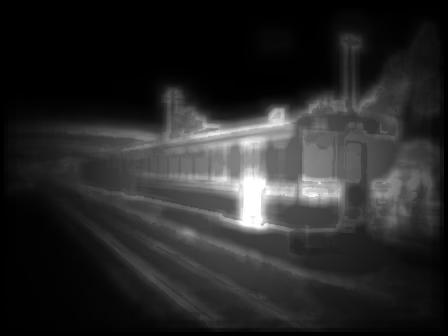}  \ &
   \includegraphics[width=0.0865\linewidth]{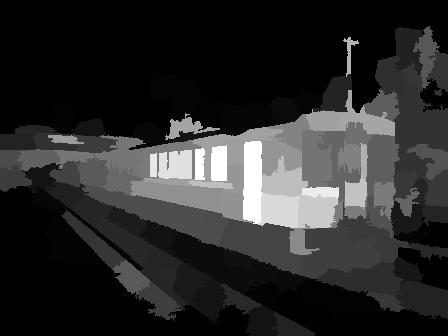}  \ &
   \includegraphics[width=0.0865\linewidth]{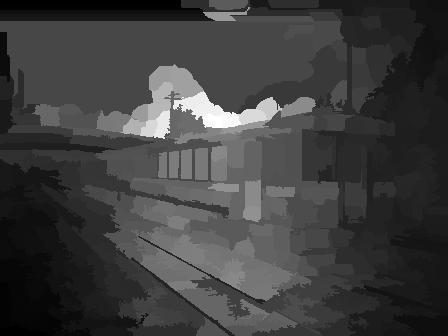}  \ &
   \includegraphics[width=0.0865\linewidth]{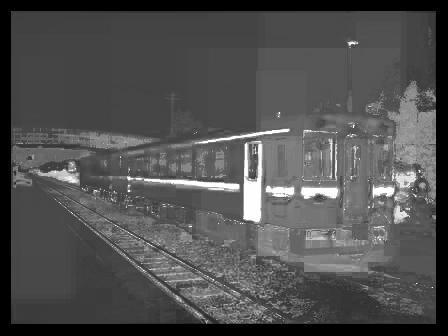}  \ &
   \includegraphics[width=0.0865\linewidth]{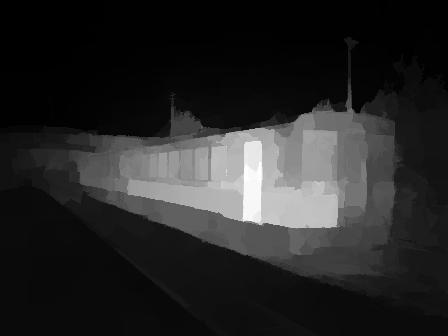}  \ \\
   \includegraphics[width=0.0865\linewidth]{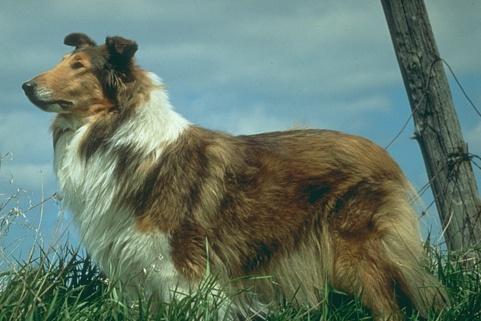}  \ &
   \includegraphics[width=0.0865\linewidth]{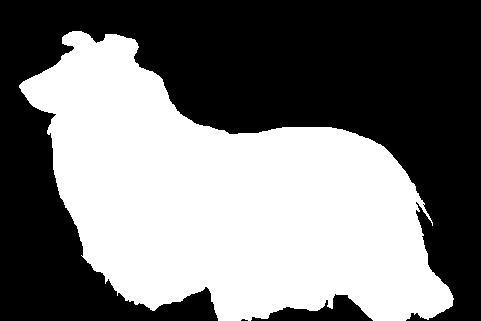}  \ &
   \includegraphics[width=0.0865\linewidth]{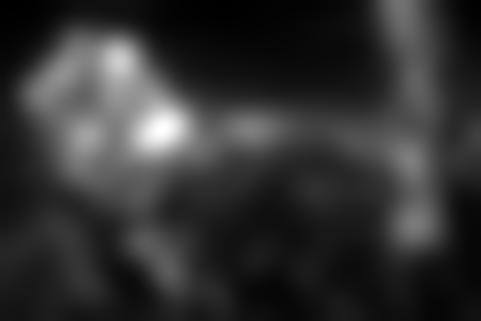}  \ &
   \includegraphics[width=0.0865\linewidth]{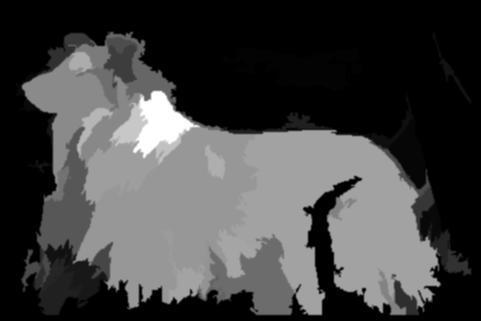}  \ &
   \includegraphics[width=0.0865\linewidth]{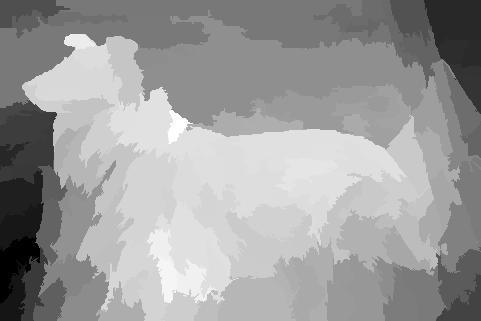}  \ &
   \includegraphics[width=0.0865\linewidth]{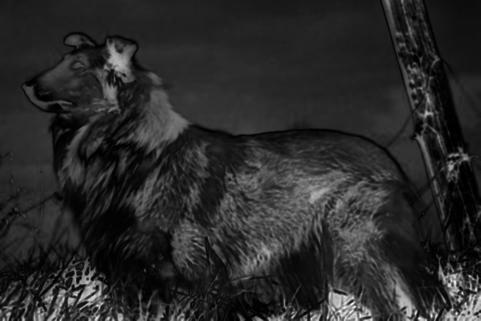}  \ &
   \includegraphics[width=0.0865\linewidth]{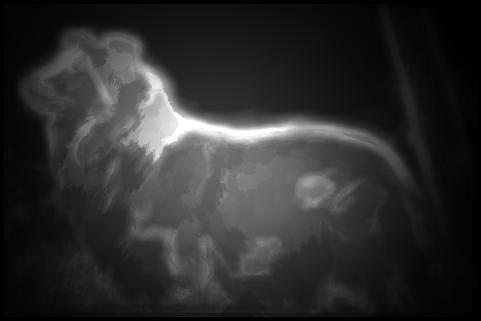}  \ &
   \includegraphics[width=0.0865\linewidth]{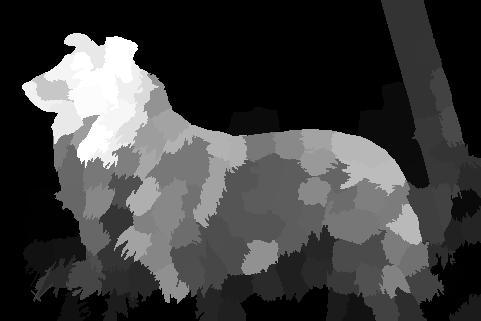}  \ &
   \includegraphics[width=0.0865\linewidth]{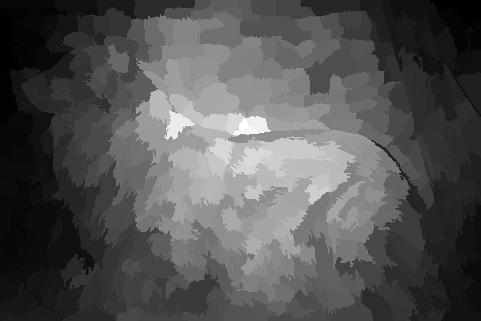}  \ &
   \includegraphics[width=0.0865\linewidth]{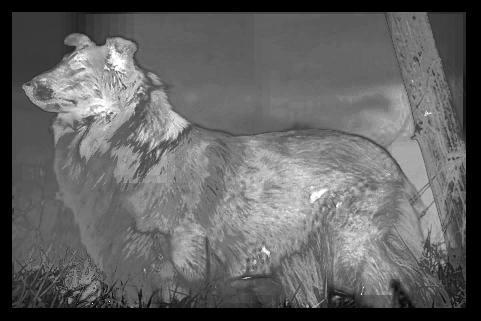}  \ &
   \includegraphics[width=0.0865\linewidth]{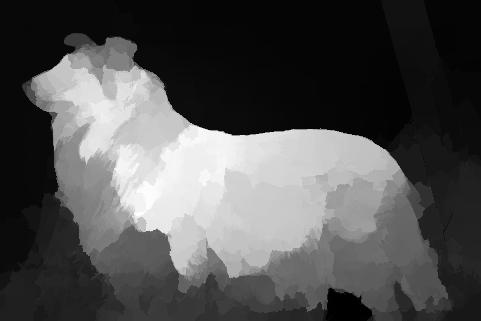}  \ \\
   {\small (a)} & {\small(b)} & {\small(c)} & {\small(d)} & {\small(e)} & {\small(f)} & {\small(g)} & {\small(h)} & {\small(i)} & {\small(j)} & {\small(k)}\ \\
\end{tabular}
\end{center}
\caption{Comparison of our saliency maps with eight state-of-the-art methods.
Left to right: (a) Test image. (b) Ground truth. (c) IT~\cite{itti1998model}. (d) RCJ~\cite{cheng2013salientPAMI}. (e) SVO~\cite{chang2011fusing}. (f) FT~\cite{achanta2009frequency}. (g) PD~\cite{margolin2013makes}. (h) GSSP~\cite{wei2012geodesic}. (i) LR~\cite{shen2012unified}. (j) RA~\cite{rahtu2010segmenting}. (k) EIS.
\label{fig:others}}
\end{figure*}
\section{Experiments}

We evaluate the proposed method on four benchmark datasets: ASD~\cite{achanta2009frequency}, THUS~\cite{cheng2013salientPAMI}, MSRA-5000~\cite{liu2011learning}, Pascal-S~\cite{li2014secrets}.
The ASD dataset is a subset of MSRA-5000, containing 1,000 images with accurate human-labelled masks.
The THUS database consists of 10,000 images, with each image having an unambiguous salient object with pixel-wise ground truth.
The MSRA-5000 dataset, which includes 5,000 more comprehensive images, has been widely used in previous saliency detection approaches.
The Pascal-S dataset is composed of 850 natural images with multiple objects and complex backgrounds.
\subsection{Evaluation of Saliency Maps}

In this paper, we generate three saliency maps including the internal saliency map (IS), external saliency map (ES) and final combined saliency map (EIS) for visual saliency detection. To demonstrate the efficiency of these maps, we select some sample results as shown in Figure~\ref{fig:first}.
The IS can separate salient regions from backgrounds in most cases, but it fails to highlight the whole object.
In contrast, the ES can detect the whole object accurately, but it sometimes brightens the background.
To overcome their shortcomings, the EIS is constructed by a weighted combination of the IS and ES, and achieves good performances on different datasets.
We also provide the PR curves of the above three maps on the MSRA-5000 and Pascal-S datasets in Figure~\ref{fig:diffmodel}.
The fused result, EIS, is apparently better than the IS and ES, which demonstrates that combining these two maps does indeed work well.
We should mention that the ES does not always outperform the IS, since it relies heavily on the image retrieval results.
Overall, the IS and ES can both highlight salient objects with great accuracy, and the performance after taking a weighted sum is superior.

\subsection{Quantitative Comparisons}
We compare the proposed saliency detection model, EIS, with 21 state-of-the-art methods including
AMC~\cite{jiang2013saliencyours}, CA~\cite{goferman2012context}, CB~\cite{jiang2011automatic}, FT~\cite{achanta2009frequency}, GB~\cite{harel2006graph}, GC~\cite{cheng2013efficient}, GSSP~\cite{wei2012geodesic}, HC~\cite{cheng2011salient}, HS~\cite{yan2013hierarchical}, IT~\cite{itti1998model}, LC~\cite{zhai2006visual}, LR~\cite{shen2012unified}, PD~\cite{margolin2013makes}, RA~\cite{rahtu2010segmenting}, RCJ~\cite{cheng2013salientPAMI}, SF~\cite{perazzi2012saliency}, SR~\cite{hou2007saliency}, SVO~\cite{chang2011fusing}, UFO~\cite{jiang2013salient}, XL~\cite{xie2013bayesian} and wCO~\cite{zhu2014saliency}.
We either use source code provided by the authors or implement them based on available code or software.

We conduct several quantitative comparisons of our EIS with some typical saliency detection approaches in this part.
Figure~\ref{fig:pr1} show the PR curves of different methods on the ASD and THUS datasets.
Figure~\ref{fig:pr2} illustrates the comparisons on the MSRA-5000 and Pascal-S datasets,
Figure~\ref{fig:bar} are relevant average precisions, recalls, F-Measures and AUCs on four datasets.
The precision and recall are computed by segmenting a saliency map with a set of thresholds varying from 0 to 255,
and comparing each binary map with the benchmark.
Our method performs well on precision-recall curves.
The highest precision rates on these four datasets are 98.2$\% $, 95.8$\% $, 93.1$\% $, and 79.9$\% $ respectively.

In addition, we evaluate the quality of saliency maps using the F-Measure and AUC.
By setting an adaptive threshold that is twice the mean saliency value of the input map, each image is segmented to a binary map.
We compute the average precision and recall based on these binary maps and computed the F-Measure as follows:
\begin{equation}\label{twnetythree}
    {F_\Upsilon } = \frac{{(1 + {\Upsilon ^2}) \times Precision \times Recall}}{{{\Upsilon ^2} \times Precision + Recall}} ,
\end{equation}
where ${{\Upsilon ^2}}$ is set to 0.3 to emphasize the precision.
Our method is comparable with most of the saliency detection approaches in terms of the F-Measure.
We also show the comparison results of AUC, which reflects global properties by computing the area under the PR curve.
Various evaluation methods on different datasets demonstrate that the proposed EIS performs favorably against the state-of-the-arts.
\subsection{Qualitative Comparisons}

Figure~\ref{fig:others} shows some example results of eight previous approaches and our EIS algorithm for qualitative comparisons.
The IT and PD methods can find salient regions in most cases, but they tend to highlight object boundaries and lose the object information.
The SVO and RA methods generate blurry saliency maps and highlight the background.
FT is easily affected by high-frequency noise and it fails to detect salient objects in all of these examples.
LR cannot highlight all the salient pixels and in all these cases mislabels small background patches as salient regions.
RCJ and GSSP are capable of finding salient regions, but they are less convincing in dealing with challenging scenes.
In constrast, our method can locate salient regions with great accuracy and highlight the whole object uniformly with unambiguous boundaries.
Furthermore, we can detect more than one object without worrying about their size and location.
\section{Conclusion}

In this paper, we propose a novel saliency detection algorithm based on the image retrieval framework.
The image retrieval framework first searches for similar examples from a subset of the CLS-LOC database to train a customized SVM for each test image, then predicts saliency values of object proposals to generate an external saliency map.
Since some images with uncommon objects may not have similar examples, we also propose an internal optimization module, which explores the contrast information within the test image by jointly optimizing the superpixel prior, discriminability, and similarity, to assist the image retrieval.
The final saliency map is generated by taking a linear combinaation of the above two maps.
We compare the proposed method with 21 state-of-the-art saliency detection approaches and show the results of precision-recall curves, average precisions, recalls, F-Measures and AUCs on four databases, including ASD, THUS, MSRA-5000, Pascal-S.
Various results demonstrate the effectiveness and efficiency of our algorithm.
In the future, we plan to design more robust image retrieval approaches to further improve the performance of our method.

{\small
\bibliographystyle{ieee}
\bibliography{egbib}
}

\end{document}